\documentclass[11pt]{article}

\usepackage[final]{acl} 
\usepackage[T1]{fontenc}
\usepackage[utf8]{inputenc}
\usepackage{microtype}
\usepackage{inconsolata}
\usepackage{times}
\usepackage{latexsym}

\usepackage{booktabs}
\usepackage{tabularx}
\usepackage{multirow}
\usepackage{makecell}
\usepackage{array}
\usepackage{amsmath, amssymb, amsthm}

\usepackage[table]{xcolor} 
\usepackage{graphicx}
\usepackage{pifont}
\usepackage{enumitem}
\usepackage{tcolorbox}
\usepackage{eso-pic} 

\definecolor{HeaderGray}{gray}{0.92}
\definecolor{LightGray}{gray}{0.95}
\definecolor{HighlightBlue}{RGB}{234, 242, 248}
\definecolor{NavyBlue}{RGB}{31, 73, 125}
\definecolor{CrimsonRed}{RGB}{192, 57, 43}
\definecolor{ForestGreen}{RGB}{39, 174, 96}
\definecolor{AxisPurple}{RGB}{112, 48, 160}
\definecolor{ErrorRed}{RGB}{192, 0, 0}
\definecolor{PassGreen}{RGB}{0, 128, 0}
\definecolor{IdxBlue}{HTML}{00509E}
\definecolor{TagTeal}{HTML}{008080}
\definecolor{WarnRed}{HTML}{D72638}
\definecolor{GtGreen}{HTML}{1A8A34}
\definecolor{FmtOrange}{HTML}{D95F02}

\providecommand{\mycmark}{\textcolor{PassGreen}{\ding{51}}}
\providecommand{\myxmark}{\textcolor{ErrorRed}{\ding{55}}}
\makeatletter
\newcommand\blfootnote[1]{%
  \begingroup
  \renewcommand\thefootnote{}\footnote{#1}%
  \addtocounter{footnote}{-1}%
  \endgroup
}
\makeatother
\AddToShipoutPictureBG*{
  \AtPageUpperLeft{
    \put(30,-60){\includegraphics[height=1.2cm]{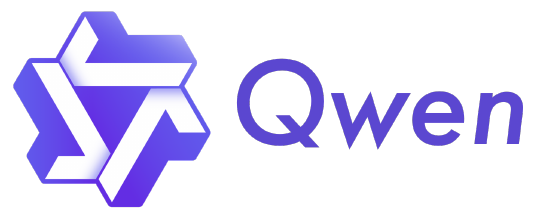}}%
  }
}

\title{Are the Financial Reasoning from LLMs Credible? A Real World Test over Long-Horizon Statements}

\author{%
 \large
 \textbf{Xinke Tong$^{1,2}$}, \textbf{Xuanming Zhang$^{1,*}$}, \textbf{Tianyi Tang$^{1}$}, \textbf{An Yang$^{1}$}, \textbf{Jiatu Hu$^{3}$}, \textbf{Guojie Lin$^{3}$}, \textbf{Zhenzhen Shi$^{3}$,} \\
 \large
 \textbf{Lingfeng Zeng$^{3}$}, \textbf{Boyu Yang$^{3}$}, \textbf{Bing Zhao$^{3}$}, \textbf{Hu Wei$^{3}$}, \textbf{Lin Qu$^{3}$}, \textbf{Dayiheng Liu$^{1}$} \\
 \vspace{0.15cm}
 \normalsize
 $^1$ Qwen Team, Alibaba Group \quad $^2$ Tsinghua University \quad $^3$ Alibaba Group \\
 \vspace{0.05cm}
 \small
 \url{https://huggingface.co/datasets/Findata/Finindice}
}

\begin{document}
\maketitle

\thispagestyle{plain} 
\pagestyle{plain}     
\blfootnote{%
  $^*$ Corresponding author. \\
  Emails: \texttt{tongxk.24@pbcsf.tsinghua.edu.cn}, \texttt{xuemuqiangu@gmail.com}%
}
\begin{abstract}
As Large Language Models (LLMs) advance, a fundamental question persists: do they possess genuine structural reasoning capabilities, or are they merely relying on surface-level pattern matching? The financial domain, with its strict demand for numerical precision, complex taxonomy, and multi-step logic over long contexts, serves as an ideal testbed for this inquiry. However, existing financial benchmarks fail to capture this real-world complexity. They predominantly rely on simplified knowledge-based multiple-choice questions or single-hop QA over artificially cropped, short-context tables, largely ignoring the intricate cross-statement dynamics, temporal de-cumulation, and specific accounting taxonomies (e.g., Chinese Accounting Standards) required in industrial workflows. 

To bridge this gap, we introduce \textsc{FinIndices}, a large-scale benchmark evaluating data-processing fidelity over uncropped, full-length financial statements (up to 32K tokens). Utilizing an automated data synthesis pipeline with adversarial traps, \textsc{FinIndices} encompasses two paradigms---\textit{Single-Index} computation and \textit{Table-Index} tabulation---to test complex domain, temporal, and caliber reasoning. 

Our comprehensive evaluation reveals two severe vulnerabilities in modern LLMs. First, a \textbf{``Knowledge Bottleneck''}: despite memorizing formulas during pre-training, models demonstrate fragile, surface-level pattern matching rather than deep comprehension. When explicit formula hints are removed, performance collapses (e.g., \textsf{Gemini-3.1-Pro} plunges from $70.70\%$ to $38.22\%$ on table tasks), exposing fatal flaws in temporal de-cumulation, stock-flow caliber mismatch, and semantic anchor traps. Second, a \textbf{``Structural Bottleneck''}: rather than mere formatting failures, the intense cognitive load of generating multi-metric, multi-period tables actively drains models' reasoning capacity. Under structural pressure, LLMs that flawlessly execute derivations in isolation systematically regress to shallow heuristics---such as fetching adjacent incorrect columns (temporal misalignment) or substituting deep accounting adjustments with lazy literal arithmetic (aggregation shortcuts). Finally, Supervised Fine-Tuning (SFT) yields substantial zero-hint gains ($+8.54\%$ Single, $+3.82\%$ Table), validating that structured logic can be partially restored via data-centric alignment.
\end{abstract}

\section{Introduction}
The rapid advancement of Large Language Models (LLMs) has catalyzed their integration into the financial domain. Recently, the field has gradually progressed toward end-to-end (E2E) agentic workflows, where models are expected to autonomously execute historical investigations, process large datasets, and generate comprehensive business reports \citep{islamg, mateega, kamble, bigeard, hu}. Accompanying this trend, an increasing number of domain-specific financial LLMs have been released (e.g., \citealt{lid, liub, zhenga}).

\begin{figure*}[t]
  \centering
  \makebox[\textwidth]{\includegraphics[width=1\textwidth]{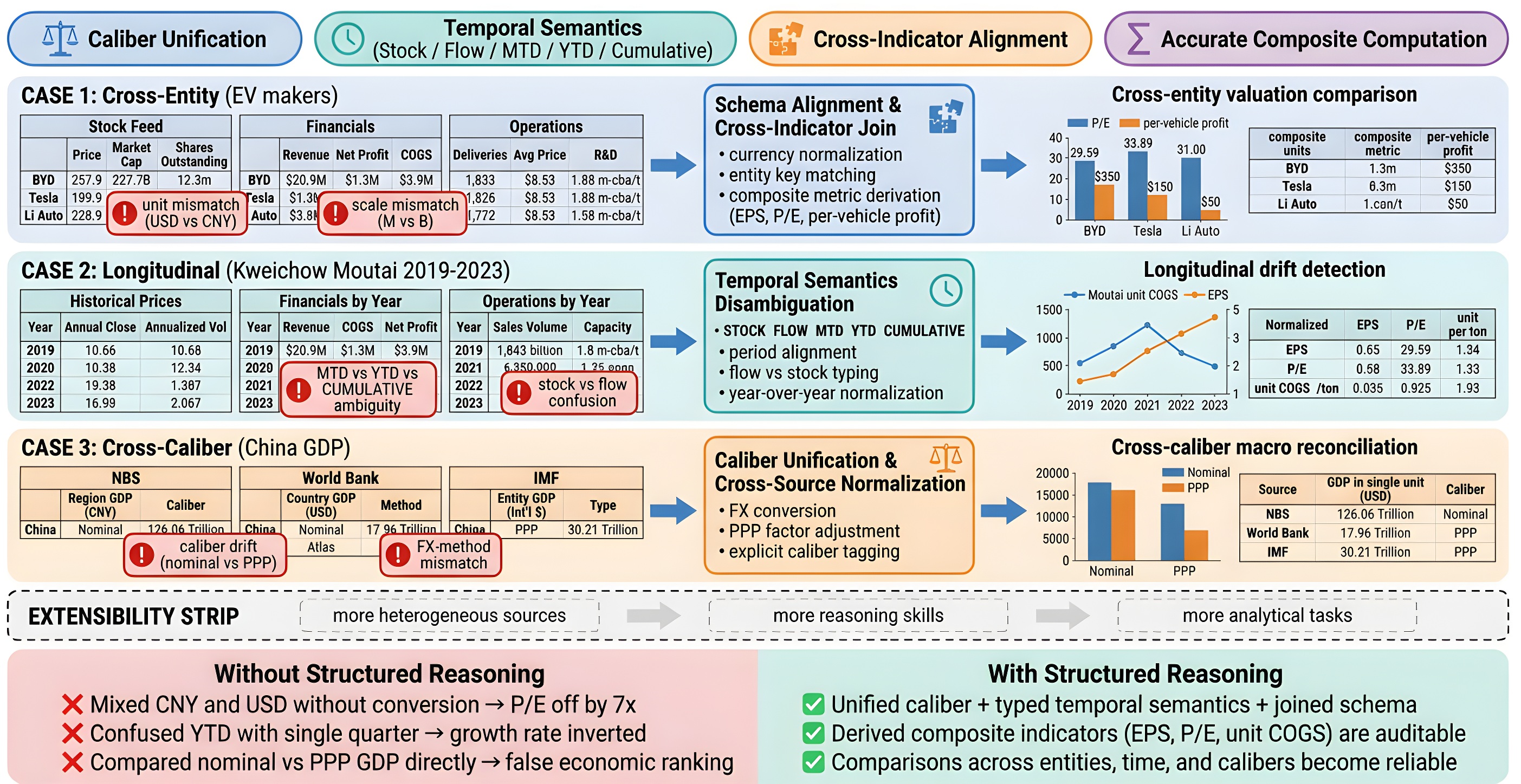}}
  \caption{
    \textbf{Three canonical financial analysis workflows motivating \textsc{FinIndices}.}
    Each case follows an \emph{input tables $\rightarrow$ structured reasoning module
    $\rightarrow$ tabular output} pipeline:
    cross-entity valuation comparison (BYD / Tesla / Li~Auto),
    longitudinal drift detection (Kweichow Moutai, 2019--2023),
    and cross-caliber macro reconciliation (China GDP under NBS / World Bank / IMF--PPP).
    Together they exemplify the three core challenges our benchmark targets: caliber
    unification, temporal-semantic disambiguation (\textsc{stock}/\textsc{flow},
    \textsc{mtd}/\textsc{ytd}/\textsc{ttm}), and cross-indicator alignment with
    accurate composite computation. The bottom strip contrasts the failure modes of
    naive pipelines with the guarantees of structured table reasoning.
  }
  \label{fig:motivation}
\end{figure*}

\paragraph{Motivation: data-processing reliability is the bottleneck of financial agents.}
A typical financial analysis task---as illustrated in Figure~\ref{fig:motivation}---rarely reduces to a single closed-form question. Instead, an analyst must (i) collect heterogeneous data from multiple sources (market feeds, financial statements, operating disclosures, macroeconomic databases), (ii) reconcile them under \emph{different statistical calibers} (e.g., parent-company vs.\ consolidated, nominal vs.\ PPP, stock vs.\ flow), \emph{different reporting frequencies} (monthly, quarterly, year-to-date, trailing twelve months), and \emph{different units and currencies}, (iii) reorganize the aligned data into well-formatted comparison tables, and (iv) derive composite indicators (e.g., EPS, P/E, unit COGS, free cash flow) for cross-entity, longitudinal, or cross-caliber analysis. The final deliverable is typically a clean, audit-ready tabular report, not a single number.

This workflow exposes a notable weakness of current financial agents. We observe that E2E pipelines often degrade not because of flawed high-level reasoning, but because of cascading errors in the intermediate stages of data extraction, alignment, and calculation. The root causes are largely \emph{domain-specific}: (1) insufficient understanding and flexible use of accounting and finance knowledge (e.g., which line items participate in a given indicator, how to handle minority interests, whether an item belongs to operating or investing cash flow); (2) fragile reasoning over \emph{complex temporal semantics} such as TTM, cumulative-to-date, single-quarter, year-over-year, and period-end vs.\ period-average values; and (3) confusion between \emph{statistical calibers} such as stock vs.\ flow, nominal vs.\ real, parent-only vs.\ consolidated, or NBS vs.\ World Bank vs.\ IMF--PPP macro statistics. When these aspects are mishandled, numbers produced by the agent become untrustworthy, blocking deployment in real business settings, where every figure in the final table must be verifiable.

\paragraph{Existing benchmarks.}
The evaluation of LLMs in the financial domain has evolved alongside model capabilities, transitioning from basic textual tasks to complex Vision--Language (VL) and agentic workflows. \emph{Early-stage benchmarks} adapted traditional NLP tasks such as sentiment analysis, named entity recognition, and text classification to finance; while foundational, they treated financial documents as general text and assessed surface-level pattern recognition rather than the numerical dependencies required in high-stakes practice. \emph{Academic-knowledge and fundamental-reasoning benchmarks} subsequently shifted to objective, knowledge-driven tasks. One stream embeds financial queries within broad benchmarks such as MMLU \citep{gemab}, MMLU-Pro \citep{wangd}, SuperGPQA \citep{team}, and PRBench \citep{akyurek}; another stream focuses exclusively on finance, including FinEval \citep{guoag}, CFinBench \citep{niec}, and FinEval-KR \citep{doub} for professional certification and economic knowledge, FinanceMATH \citep{zhaobi} and FinanceReasoning \citep{tang} for multi-step mathematical reasoning, and FinBen \citep{xiet} and FinTagging \citep{wangix} for academic competencies such as XBRL taxonomy linking.

To better reflect professional workflows, a third line of \emph{industrial-skill benchmarks} has emerged. Initial efforts leveraged public filings for complex QA over hybrid textual--tabular data \citep{chenfj, zhu}, later extended to open-book QA \citep{islamg}, incomplete-information analysis \citep{mateega}, and robustness testing against query perturbations \citep{kamble}. More recent benchmarks such as BizFinBench \citep{lu}, FLAME \citep{guoaf}, and FIRE \citep{zhangd} incorporate proprietary business logic and precision-critical scenarios, evaluating models on practical applications via expert-defined rubrics. In parallel, with advancing multimodal capabilities, \emph{Vision--Language financial benchmarks} such as MME-Finance \citep{gane}, FAMMA \citep{xued}, and FinMME \citep{luo} use multiple-choice and precise-calculation tasks to evaluate both academic-style visual reasoning (e.g., interpreting tabular structures and formulas) and practical applications (e.g., analyzing stock K-line charts and research infographics). Finally, \emph{agentic benchmarks} including Finance Agent Benchmark \citep{bigeard} and FinSearchComp \citep{hu} move beyond static QA, requiring models to interact with external tools (e.g., web search, SEC EDGAR, Python execution) for dynamic information gathering and multi-step historical investigations that mimic real analyst workflows.

\paragraph{Gaps in existing benchmarks.}
Despite this rich landscape, existing benchmarks do not adequately measure the intermediate \emph{data-processing reliability} that dominates real financial workflows. Roughly, prior tabular-oriented evaluations fall into two camps. The first focuses on \emph{simple tabular reasoning with one or a few numerical answers}, such as FinQA \citep{chenfj}, TAT-QA \citep{zhu}, and the practical-skill subsets of FLAME \citep{guoaf} and FIRE \citep{zhangd}. These datasets typically operate on localized, pre-cropped table snippets, involve relatively shallow operators, and are often programmatically synthesized; they rarely test the temporal and caliber-related concepts that dominate real analyses, nor do they require producing structured multi-row, multi-column outputs. The second camp is more \emph{academically oriented}, consisting of knowledge-style multiple-choice questions drawn from professional examinations or textbooks (e.g., portions of FinEval \citep{guoag}, CFinBench \citep{niec}, FinEval-KR \citep{doub}, FinBen \citep{xiet}, and BizFinBench \citep{lu}). While useful for probing factual knowledge, such formats are far from the data-intensive, table-centric workflows that financial practitioners actually perform. Multimodal benchmarks \citep{gane, xued, luo} and agentic benchmarks \citep{bigeard, hu} likewise emphasize either visual perception or end-to-end tool use, leaving the underlying numerical-table reliability largely unmeasured. Moreover, comprehensive Chinese-language tabular QA resources targeting these aspects remain scarce, since existing Chinese financial evaluations include tabular QA only as a minor subset.

\paragraph{Our benchmark.}
To close this gap, we introduce \textbf{FinIndices}, a benchmark specifically designed to evaluate \textbf{calculation fidelity and data-processing reliability} in realistic financial workflows. We collect continuous, fine-grained financial statements spanning multiple reporting frequencies and statistical calibers, together with a curated set of indicators grounded in canonical textbooks and accounting standards. Built on top of these resources, FinIndices is organized along \textbf{two task formats}---\emph{single-value computation} and \emph{unified-caliber multi-indicator tabulation}---and \textbf{three orthogonal capability axes}: \emph{complex accounting-domain understanding} (e.g., free cash flow, EBITDA), \emph{complex temporal reasoning} (e.g., single-quarter QoQ, TTM), and \emph{complex caliber alignment} (e.g., stock--flow unification for inventory turnover). By evaluating models along these dimensions, FinIndices serves as a \textbf{practical testbed for identifying intermediate failure modes} and improving the data reliability of future financial LLMs and agentic pipelines.

\section{Method}
\subsection{Data Synthesis}

\begin{figure*}[t]
  \centering
  \makebox[\textwidth]{\includegraphics[width=1\textwidth]{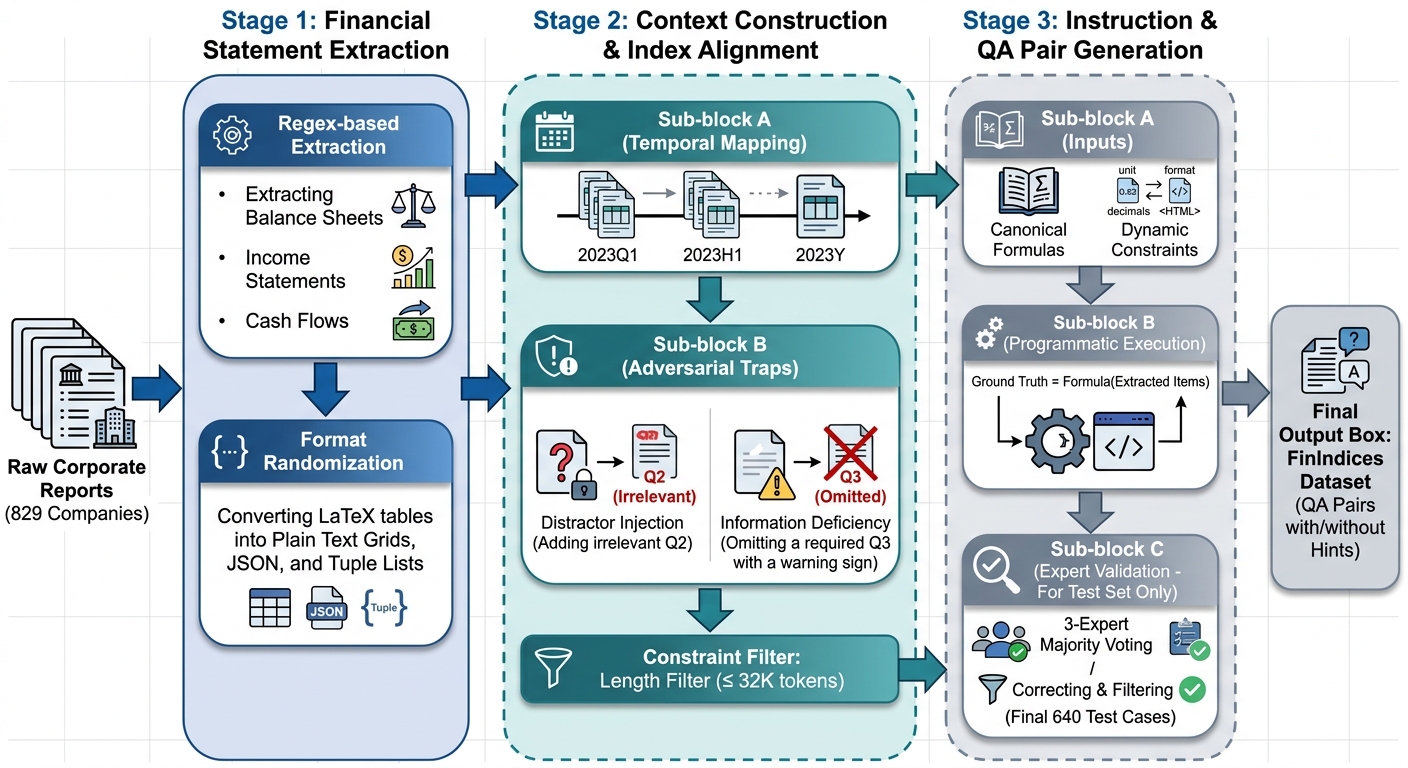}}
  \caption{\textbf{Overview of the \textsc{FinIndices} data synthesis pipeline.} The workflow consists of three stages: (1) \textbf{Financial Statement Extraction} parses and format-randomizes core tables from raw reports; (2) \textbf{Context Construction \& Index Alignment} aligns temporal data and injects adversarial traps (e.g., missing info); and (3) \textbf{Instruction \& QA Pair Generation} constructs prompts using canonical formulas and programmatic execution, followed by a rigorous \textbf{Expert Validation} phase employing majority voting to ensure absolute ground-truth fidelity.}
  \label{fig:data_synthesis}
\end{figure*}

To construct the dataset, we develop an automated, scalable data synthesis pipeline that extracts, aligns, and generates question-answering pairs from authentic corporate financial reports. The synthesis process covers 829 publicly listed companies, 384 diverse financial indices, and spans 28 reporting periods, ensuring comprehensive temporal and cross-sectional coverage. The pipeline is structured into three granular stages: Financial Statement Extraction, Context Construction \& Index Alignment, and Instruction \& QA Pair Generation.

\paragraph{Financial Statement Extraction.}
The foundation of our dataset relies on accurately extracting core financial tables (i.e., Balance Sheet, Income Statement, and Cash Flow Statement) from raw corporate reports. We employ regular expression-based heuristics to isolate the ``Financial Statements'' section from surrounding narrative texts. To enhance the model's structural generalization capabilities, the parsed \LaTeX-formatted tabular blocks (\verb|\begin{tabular}...\end{tabular}|) are randomly converted into diverse representations, including Plain Text grids, JSON, and Tuple Lists. Finally, basic filtering mechanisms are applied to discard malformed or incomplete tables, ensuring high data fidelity.

\paragraph{Context Construction \& Index Alignment.}
Financial indices are intrinsically temporal, often requiring data across multiple periods (e.g., Year-over-Year (YoY), Quarter-over-Quarter (QoQ), or Trailing Twelve Months (TTM)). We construct rigorous contexts based on query types:
\begin{itemize}
    \item \textbf{Temporal Mapping for Single-Index Queries:} We retrieve the exact combination of reporting periods needed for calculation (e.g., current quarter and the quarter from the previous year for YoY). 
    \item \textbf{Adversarial Trap Injection:} To evaluate true comprehension rather than blind computation, we design two types of traps: 
    (1) \textit{Distractor Injection:} Randomly inserting adjacent but irrelevant periods (e.g., adding Q1 when the query only requires Q2 and Q3). 
    (2) \textit{Information Deficiency:} Deliberately omitting necessary preceding reports, forcing the model to explicitly output ``Insufficient information'' rather than hallucinating a numeric value.
    \item \textbf{Multi-Period Aggregation for Table-Index Queries:} For tasks requiring time-series outputs, we dynamically sample consecutive periods (ranging from 4 to 9 periods). The corresponding financial tables are concatenated into a unified text block.
    \item \textbf{Length Constraint:} To closely align with real-world business application scenarios, we strictly filter out contexts exceeding 32,768 tokens using a pre-defined tokenizer.
\end{itemize}

\paragraph{Instruction \& QA Pair Generation.}
To simulate diverse and naturalistic user interactions, we utilize a template-based instruction generation module equipped with dynamic constraints and rigorous validation:
\begin{itemize}
    \item \textbf{Canonical Formulas \& Programmatic Execution:} To guarantee absolute computational accuracy, the calculation formulas (i.e., ``hints'') injected into the prompts are curated strictly from canonical financial textbooks and formal accounting standards, rather than third-party web platforms. The initial ground-truth answers are derived via programmatic execution: our system extracts the exact source line items required by the formula from the corresponding financial statements and executes the arithmetic. 
    \item \textbf{Diverse Query Templating:} We sample from a rich pool of human-written question templates (e.g., ``Calculate the [Index] of [Company] for [Period]'', ``What is the value of...''). 
    \item \textbf{Dynamic Constraints:} We randomly inject computational and formatting constraints into the prompts. This includes unit conversions (e.g., converting absolute values into ``Yuan'', ``Thousands'', ``Millions'', or ``Billions'') and numerical precision requirements (e.g., ``round to 1--3 decimal places''). The programmatically executed ground-truth answers are dynamically scaled and rounded to perfectly match these injected constraints.
    \item \textbf{Structured Output Formatting:} For multi-index or multi-period queries (Table-Index), the instructions explicitly require the model to output the final results in specified machine-readable formats (e.g., encapsulated within \verb|<table>...</table>| HTML tags or \verb|```json ... ```| blocks).
    \item \textbf{Expert Validation \& Test Set Refinement:} To ensure the highest evaluation standard, the initial test split (684 instances) underwent rigorous auditing by domain experts. During validation, experts not only verified whether the input context was sufficient to compute the target value but also audited the numerical accuracy of the programmatic answers, manually correcting any errors or rounding discrepancies. Instances where derivation was impossible due to structural missing data were annotated as ``Insufficient information.'' We employed a majority-voting mechanism among three independent experts: 484 instances reached full consensus, and 171 instances reached a two-person majority, totaling 655 verified instances. Finally, to ensure a balanced sample distribution and prevent a skew toward unanswerable queries, we pruned a subset of the ``Insufficient information'' samples, yielding a final, ultra-reliable test set of 640 instances.
\end{itemize}

\subsection{Dataset Analysis}
\subsubsection{Task Description and Statistics}

\definecolor{AxisPurple}{HTML}{6A1B9A}

To comprehensively evaluate data-processing reliability, \textsc{FinIndices} is structured along two orthogonal dimensions: the \textbf{Task Paradigm} (output structural complexity) and the \textbf{Capability Axis} (core financial reasoning skill). Table~\ref{tab:capability_distribution} details the dataset partitioning and breaks down the capability distribution. Detailed worked examples for these dimensions are provided in Appendix~\ref{sec:appendix_case_single_domain} to \ref{sec:appendix_case_table_temporal_caliber}.

\begin{table*}[t]
\centering
\footnotesize 
\renewcommand{\arraystretch}{1.15} 

\begin{tabular*}{\textwidth}{@{\extracolsep{\fill}} l l rr rr rr rr r @{}}
\toprule
\multirow{2}{*}{\textbf{Split}} & \multirow{2}{*}{\textbf{Query Paradigm}}
& \multicolumn{2}{c}{\textcolor{AxisPurple}{\textbf{Domain}}}
& \multicolumn{2}{c}{\textcolor{AxisPurple}{\textbf{Caliber}}}
& \multicolumn{2}{c}{\textcolor{AxisPurple}{\textbf{Temporal} $+$ \textbf{Caliber}}}
& \multicolumn{2}{c}{\textcolor{AxisPurple}{\textbf{Temporal} $+$ \textbf{Domain}}}
& \multirow{2}{*}{\textbf{Total}} \\
\cmidrule(lr){3-4} \cmidrule(lr){5-6} \cmidrule(lr){7-8} \cmidrule(lr){9-10}
& & \# & \% & \# & \% & \# & \% & \# & \% & \\
\midrule

\multirow{3}{*}{\textbf{Train}}
& Single-Index & 4{,}192 & 7.27 & 36{,}829 & 63.85 & 10{,}015 & 17.36 & 6{,}645 & 11.52 & 57{,}681 \\
& Table-Index  & 1{,}131 & 11.75 &  5{,}579 & 57.96 &  1{,}735 & 18.02 & 1{,}181 & 12.27 &  9{,}626 \\
\cmidrule{2-11}
& \textbf{Subtotal} & \textbf{5{,}323} & \textbf{7.91} & \textbf{42{,}408} & \textbf{63.01} & \textbf{11{,}750} & \textbf{17.46} & \textbf{7{,}826} & \textbf{11.63} & \textbf{67{,}307} \\

\midrule

\multirow{3}{*}{\textbf{Dev}}
& Single-Index & 15 &  6.91 & 138 & 63.59 & 38 & 17.51 & 26 & 11.98 & 217 \\
& Table-Index  & 14 &  9.79 &  89 & 62.24 & 21 & 14.69 & 19 & 13.29 & 143 \\
\cmidrule{2-11}
& \textbf{Subtotal} & \textbf{29} & \textbf{8.06} & \textbf{227} & \textbf{63.06} & \textbf{59} & \textbf{16.39} & \textbf{45} & \textbf{12.50} & \textbf{360} \\

\midrule

\multirow{3}{*}{\textbf{Test}}
& Single-Index & 34 &  7.04 & 311 & 64.39 & 82 & 16.98 & 56 & 11.59 & 483 \\
& Table-Index  & 22 & 14.01 &  86 & 54.78 & 32 & 20.38 & 17 & 10.83 & 157 \\
\cmidrule{2-11}
& \textbf{Subtotal} & \textbf{56} & \textbf{8.75} & \textbf{397} & \textbf{62.03} & \textbf{114} & \textbf{17.81} & \textbf{73} & \textbf{11.41} & \textbf{640} \\

\bottomrule
\end{tabular*}
\caption{\textbf{Distribution of \textsc{FinIndices} along the capability axes.} Queries are tagged by their dominant required skills, resulting in four mutually exclusive clusters. Proportions remain stable across all data splits.}
\label{tab:capability_distribution}
\end{table*}

\begin{table*}[t]
\centering
\small 
\renewcommand{\arraystretch}{1.15} 
\setlength{\tabcolsep}{16pt} 

\begin{tabular}{l r rrr}
\toprule
\multirow{2}{*}{\textbf{Dataset}} & \multirow{2}{*}{\textbf{Samples}} & \multicolumn{3}{c}{\textbf{Context Length (Tokens)}} \\
\cmidrule(lr){3-5}
& & \textbf{Mean} & \textbf{Max} & \textbf{Min} \\
\midrule
FinQA \citep{chenfj}  &  8,281 &   1,003.8 &  3,250 & 168 \\
TAT-QA \citep{zhu}    & 14,883 &     590.7 &  4,140 & 113 \\
\textbf{\textsc{FinIndices} (Ours)} & \textbf{68,307} & \textbf{16,202.0} & \textbf{33,126} & \textbf{70} \\
\midrule[0.8pt]
\multicolumn{5}{l}{\textit{Temporal and Structural Complexity in \textsc{FinIndices}}} \\
\midrule
\textbf{Query Paradigm} & \textbf{Input Periods} & \textbf{Output Periods} & \multicolumn{2}{c}{\textbf{Target Values Generated}} \\
\midrule
Single-Index Queries & Avg: 1.95 (Max: 6) & Avg: 1.00 (Max: 1) & \multicolumn{2}{c}{1.00 (Single value)} \\
Table-Index Queries  & Avg: 2.89 (Max: 9) & Avg: 2.03 (Max: 4) & \multicolumn{2}{c}{Avg: 7.75 (Range: 2 -- 20)} \\
\bottomrule
\end{tabular}
\caption{\textbf{Context Length and Structural Complexity.} \textsc{FinIndices} features an order-of-magnitude increase in context lengths compared to previous datasets. Furthermore, to resolve these queries, models must aggregate data across multiple interconnected reporting periods and, in Table-Index tasks, generate highly structured multi-value outputs.}
\label{tab:context_complexity}
\end{table*}

\paragraph{Dimension I: Task Paradigms.} 
To simulate varying structural demands, we define two primary output formats alongside an adversarial check:
\begin{itemize}
    \item \textbf{Single-Value Computation (Single-Index):} Models must calculate one numerical answer from long, distractor-rich inputs, isolating calculation fidelity (Appendix~\ref{sec:appendix_case_single_domain}).
    \item \textbf{Unified-Caliber Multi-Indicator Tabulation (Table-Index):} Models must compute multiple interrelated metrics and assemble them into strictly formatted HTML or JSON structures without generation collapse (Appendix~\ref{sec:appendix_case_table_domain}).
    \item \textbf{Adversarial Queries:} Critical reports are deliberately omitted from the context, forcing models to explicitly reject the query rather than hallucinate numbers (Appendix~\ref{sec:appendix_case_single_temporal_adv}).
\end{itemize}

\paragraph{Dimension II: Capability Axes.}
Orthogonal to the task paradigms, every query is categorized by the deep financial reasoning skills it evaluates:
\begin{itemize}
    \item \textbf{Pure Caliber Alignment (\textcolor{AxisPurple}{Caliber}):} Evaluates the capability to reconcile and align different natures of financial metrics, distinguishing correctly between point-in-time (stock) and period-aggregated (flow) data structures (Appendix~\ref{sec:appendix_case_table_caliber}).
    
    \item \textbf{Pure Accounting-Domain Understanding (\textcolor{AxisPurple}{Domain}):} Assesses the deep comprehension of complex accounting principles, financial statement hierarchies, and the precise categorization of specialized financial terminologies (Appendix~\ref{sec:appendix_case_single_domain}).
    
    \item \textbf{Temporal + Caliber Alignment (\textcolor{AxisPurple}{Temporal $+$ Caliber}):} Challenges the model to perform structural caliber conversions across varying time horizons, ensuring metrics remain logically consistent when navigating multi-period or granular data formats (Appendix~\ref{sec:appendix_case_table_temporal_caliber}).
    
    \item \textbf{Temporal + Domain Understanding (\textcolor{AxisPurple}{Temporal $+$ Domain}):} Requires synthesizing domain-specific accounting expertise across continuous or non-sequential financial periods to compute dynamic, time-sensitive indicators (Appendix~\ref{sec:appendix_case_single_temporal_domain}).
\end{itemize}

\subsubsection{Automated Evaluation Protocol}
\label{sec:evaluation_protocol}

We introduce a rigorous automated scoring pipeline to evaluate both mathematical precision and structural compliance.

For \textit{table-index} queries, the Ground Truth (GT) is inherently a matrix $\mathbf{M} \in \mathbb{R}^{N \times T}$, where $N$ is the number of indices and $T$ represents reporting periods. We flatten this matrix into a target set of required values:
\begin{equation}
\mathcal{V}_{gt} = \{ m_{i,j} \mid 1 \le i \le N, 1 \le j \le T \}
\end{equation}

Given the raw model output $O$ and the requested format $F \in \{\text{HTML, JSON, Tuple}\}$, we apply a format parser to isolate the core structural block $B = \Psi(O, F)$. We then employ a robust regex extractor $\Phi(B)$ to obtain a multiset of predicted values $\mathcal{V}_{pred}$, effectively neutralizing financial formatting noise such as thousand-separator commas.

To evaluate correctness, we define mathematical equivalence ($\approx$) to match symbolic numeric values (e.g., $1,234.50 \approx 1234.5$). To prevent partial credit for incomplete or ill-formatted tables, we adopt a strict \textbf{all-or-nothing} scoring metric $S$:
\begin{equation}
S = \prod_{g \in \mathcal{V}_{gt}} \mathbb{1} \left( \exists d \in \mathcal{V}_{pred} \text{ s.t. } d \approx g \right)
\end{equation}
where $\mathbb{1}(\cdot)$ is the indicator function. A score of $1$ is awarded \textit{if and only if} all required metrics are accurately calculated and correctly encapsulated within the requested format tags; otherwise, $S = 0$.

\definecolor{Top1}{HTML}{D72638} 
\definecolor{Top2}{HTML}{007BFF} 
\definecolor{top1color}{HTML}{B03A2E} 
\definecolor{top2color}{HTML}{2874A6} 

\newcommand{\TopOne}[1]{\textcolor{top1color}{\textbf{#1}}}
\newcommand{\TopTwo}[1]{\textcolor{top2color}{\textbf{#1}}}
\newcommand{\std}[1]{$_{\pm #1}$} 
\newcommand{\posdelta}[1]{\textcolor{green!50!black}{\textbf{#1}}} 

\begin{table*}[t]
\centering
\setlength{\tabcolsep}{2.5pt}
\renewcommand{\arraystretch}{1.3}
\resizebox{\textwidth}{!}{
\begin{tabular}{@{} l ccc ccc c ccc ccc @{}}
\toprule

\multirow{3}{*}{\textbf{Models}} & \multicolumn{6}{c}{\textbf{With Hint (w/ Formula)}} & & \multicolumn{6}{c}{\textbf{Without Hint (w/o Formula)}} \\
\cmidrule(lr){2-7} \cmidrule(l){9-14}
& \multicolumn{3}{c}{\textbf{Single-Index}} & \multicolumn{3}{c}{\textbf{Table-Index}} & & \multicolumn{3}{c}{\textbf{Single-Index}} & \multicolumn{3}{c}{\textbf{Table-Index}} \\
\cmidrule(lr){2-4} \cmidrule(lr){5-7} \cmidrule(lr){9-11} \cmidrule(l){12-14}
& Train & Dev & \textbf{Test} & Train & Dev & \textbf{Test} & & Train & Dev & \textbf{Test} & Train & Dev & \textbf{Test} \\
\midrule

\rowcolor[gray]{0.9} 
\multicolumn{14}{l}{\textbf{Closed-Source Models}} \\ \midrule
\textsf{Gemini-3.1-Pro (thinking)}  & 66.29 & 69.05\std{0.87} & \TopOne{79.61\std{1.05}} & 34.50 & 26.57\std{1.51} & \TopTwo{70.70\std{0.30}} & & 55.95 & 58.96\std{1.21} & \TopTwo{64.90\std{0.82}} & 17.35 & 9.09\std{0.33} & \TopTwo{38.22\std{0.52}} \\
\textsf{Claude-Opus-4.8 (thinking)} & 66.91 & 72.66\std{0.78} & 78.46\std{0.57} & 31.46 & 23.08\std{0.99} & 65.61\std{1.38} & & 56.65 & 60.62\std{1.32} & \TopOne{65.79\std{0.46}} & 16.51 &  9.79\std{0.99} & \TopOne{38.85\std{1.38}} \\
\textsf{GPT-5.5 (thinking)}         & 65.69 & 74.46\std{1.04} & 77.29\std{0.34} & 32.90 & 22.84\std{0.66} & 66.88\std{0.52} & & 54.42 & 59.17\std{0.13} & 60.91\std{1.10} & 15.12 & 11.66\std{0.33} & 34.18\std{0.30} \\
\textsf{GPT-5.4 (thinking)}         & 65.88 & 71.58\std{0.89} & 75.79\std{0.81} & 33.38 & 25.17\std{0.75} & 65.61\std{0.92} & & 55.96 & 59.02\std{0.78} & 57.04\std{0.84} & 14.77 & 10.49\std{0.41} & 31.21\std{0.62} \\
\midrule

\rowcolor[gray]{0.9} 
\multicolumn{14}{l}{\textbf{Open-Weight Large Models ( $\mathbf{\geq}$ 200B )}} \\ \midrule
\textsf{Qwen3.7-Max (thinking)}     & 62.59 & 69.51\std{0.93} & \TopTwo{78.53\std{0.28}} & 35.10 & 26.57\std{0.00} & \TopOne{72.82\std{0.60}} & & 50.75 & 54.24\std{0.67} & 58.18\std{0.92} & 15.47 & 10.02\std{0.66} & 33.97\std{0.79} \\
\textsf{DeepSeek-V4-Pro (thinking)} & 64.75 & 66.00\std{1.23} & 69.20\std{0.79} & 25.28 & 15.62\std{1.74} & 48.20\std{1.31} & & 52.85 & 50.69\std{0.32} & 52.74\std{0.38} &  9.86 &  6.29\std{0.57} & 22.72\std{2.10} \\
\textsf{GLM-5.2 (thinking)}         & 51.14 & 57.41\std{0.68} & 60.50\std{1.09} & 16.77 & 11.19\std{0.99} & 36.94\std{0.52} & & 43.17 & 45.05\std{0.13} & 48.82\std{1.12} &  7.27 &  5.59\std{0.57} & 19.53\std{1.83} \\
\textsf{GLM-5 (thinking)}           & 51.28 & 50.93\std{0.65} & 50.85\std{0.74} & 16.84 & 12.59\std{0.52} & 35.03\std{0.68} & & 41.22 & 35.21\std{0.48} & 36.97\std{0.55} &  7.98 &  5.59\std{0.28} & 19.11\std{0.42} \\
\midrule

\rowcolor[gray]{0.9} 
\multicolumn{14}{l}{\textbf{Open-Weight Medium Models ( 50B -- 200B )}} \\ \midrule
\textsf{Qwen3.7-Plus (thinking)}    & 61.53 & 65.33\std{0.69} & 76.16\std{1.11} & 30.58 & 21.91\std{0.87} & 63.91\std{0.60} & & 50.38 & 53.77\std{1.74} & 58.72\std{1.01} & 13.63 &  8.39\std{1.51} & 29.72\std{2.86} \\
\textsf{Qwen3.5-Plus (thinking)}    & 61.56 & 60.70\std{0.75} & 68.78\std{0.88} & 29.64 & 26.57\std{0.55} & 63.69\std{0.72} & & 48.91 & 54.19\std{0.68} & 50.11\std{0.74} & 12.99 &  8.39\std{0.35} & 26.75\std{0.65} \\
\midrule

\rowcolor[gray]{0.9} 
\multicolumn{14}{l}{\textbf{Open-Weight Small Models ( $<$ 50B )}} \\ \midrule
\multicolumn{14}{@{}l}{\textit{General}} \\ \midrule
\hspace{3mm}\textsf{Qwen3.5-35B-A3B-thinking} & 55.12 & 61.85\std{0.82} & 62.40\std{0.76} & 14.85 & 11.10\std{0.45} & 26.50\std{0.55} & & 42.15 & 47.10\std{0.58} & 48.20\std{0.61} &  6.55 &  7.15\std{0.31} & 12.50\std{0.38} \\
\midrule
\multicolumn{14}{@{}l}{\textit{Domain-Specific Financial}} \\ \midrule
\hspace{3mm}\textsf{DianJin-R1-32B} \citep{zhubi} & 37.94 & 45.97\std{0.81} & 41.29\std{0.75} &  5.01 &  2.90\std{0.18} & 11.41\std{0.42} & & 29.40 & 34.00\std{0.66} & 32.39\std{0.58} &  2.08 &  0.74\std{0.05} &  5.33\std{0.21} \\
\hspace{3mm}\textsf{Fin-o1-14B}                 & 24.94 & 22.89\std{0.45} & 28.16\std{0.53} &  1.12 &  0.71\std{0.08} &  2.67\std{0.15} & & 18.60 & 20.87\std{0.38} & 18.32\std{0.41} &  0.55 &  0.00\std{0.00} &  1.97\std{0.11} \\
\hspace{3mm}\textsf{XuanYuan-FinX1}             &  9.15 &  6.82\std{0.25} & 12.45\std{0.33} &  1.26 &  0.70\std{0.05} &  1.91\std{0.10} & &  5.85 &  7.12\std{0.20} &  5.33\std{0.18} &  0.51 &  0.00\std{0.00} &  0.64\std{0.04} \\
\hspace{3mm}\textsf{Fin-R1} \citep{liub}        &  8.21 &  4.95\std{0.18} & 10.87\std{0.28} &  0.08 &  0.00\std{0.00} &  0.00\std{0.00} & &  4.98 &  6.25\std{0.15} &  4.41\std{0.12} &  0.02 &  0.00\std{0.00} &  0.00\std{0.00} \\
\hspace{3mm}\textsf{Llama-Fin-8B}               &  7.92 &  4.50\std{0.12} & 10.15\std{0.22} &  0.00 &  0.00\std{0.00} &  0.00\std{0.00} & &  4.65 &  5.81\std{0.11} &  4.12\std{0.09} &  0.00 &  0.00\std{0.00} &  0.00\std{0.00} \\
\bottomrule
\end{tabular}
}
\caption{\textbf{Main Results on \textsc{FinIndices}.} Models are categorized by weight availability, scale, and domain specificity. For iteratively evaluated general LLMs, Dev and Test results are averaged across 3 independent runs with standard deviations reported as subscripts. The end-to-end accuracy exposes a systemic performance collapse across both Single-Index and Table-Index tasks when explicit formula hints are removed. Top-2 Test results are highlighted in \TopOne{Red (1st)} and \TopTwo{Blue (2nd)}.}
\label{tab:main_results}
\end{table*}

\subsubsection{Dataset Statistics and Comparison}
\label{sec:dataset_statistics}

\begin{table*}[t]
\centering
\small
\setlength{\tabcolsep}{4pt}
\renewcommand{\arraystretch}{1.15} 
\resizebox{\textwidth}{!}{
\begin{tabular}{@{} l cccc c cc @{}}
\toprule

\multirow{3}{*}{\textbf{Models}} & \multicolumn{4}{c}{\textbf{Finance Knowledge}} & \textbf{Finance Calc.} & \multicolumn{2}{c}{\textbf{Finance Table Reasoning}} \\
\cmidrule(lr){2-5} \cmidrule(lr){6-6} \cmidrule(l){7-8}
& CFinBench & FinEval & FLAME & In-House & FinMath & Single-Index & Table-Index \\
& \citep{niec} & \citep{guoag} & \citep{guoaf} &  &  & \textit{(w/o Hint)} & \textit{(w/o Hint)} \\
\midrule

\rowcolor[gray]{0.95} 
\multicolumn{8}{l}{\textit{\textbf{Ablation: Impact of Domain-Specific Fine-Tuning}}} \\ \midrule

\textsf{Qwen3.5-35B-A3B-thinking (Base)} & 74.51 & 88.59 & 84.87 & 71.95 & 66.90 & 47.92 & 15.29 \\
\textbf{Ours (SFT on Base)}              & 75.22 & 89.43 & 87.26 & 72.56 & 69.08 & 56.46 & 19.11 \\ \midrule
\textbf{Absolute Improvement ($\Delta$)} & \posdelta{+0.72} & \posdelta{+0.84} & \posdelta{+2.39} & \posdelta{+0.61} & \posdelta{+2.19} & \posdelta{+8.54} & \posdelta{+3.82} \\

\bottomrule
\end{tabular}
}
\caption{\textbf{Improvements via Domain-Specific SFT.} Our fine-tuned variant shows steady gains on general knowledge benchmarks and substantial improvements in complex numerical reasoning and tabular generation (\textit{Without Hint} setting).}
\label{tab:sft_improvement}
\end{table*}

To evaluate the capabilities of Large Language Models (LLMs) in realistic financial reasoning scenarios, we construct \textbf{\textsc{FinIndices}}, a large-scale benchmark featuring extended tabular contexts. As detailed in Table~\ref{tab:capability_distribution}, the dataset comprises 68,307 samples. We outline the key characteristics of our dataset by comparing it with existing tabular QA benchmarks across three structural dimensions, as summarized in Table~\ref{tab:context_complexity}.

\paragraph{Extended Tabular Contexts with Distractors.}
Existing datasets like FinQA \citep{chenfj} and TAT-QA \citep{zhu} typically rely on pre-cropped table snippets, resulting in average context lengths of 500–1,000 tokens. In contrast, \textsc{FinIndices} provides uncropped financial statements, bringing the average context length to over 16,000 tokens (with a maximum of 33,126 tokens). Rather than simply increasing length, this broader scope requires models to navigate a large volume of irrelevant financial metrics, thereby testing their information extraction capabilities in the presence of substantial tabular noise.

\paragraph{Complex Temporal Spans.}
Financial analysis frequently involves dynamic comparisons over time. As shown in the bottom half of Table~\ref{tab:context_complexity}, resolving a single query in \textsc{FinIndices} requires models to process and align data spanning multiple historical reporting periods (\emph{Input Periods} averaging 1.95 to 2.89, up to 9). Furthermore, the expected output targets also span multiple periods (\emph{Output Periods}), which evaluates the model’s ability to maintain temporal alignment during cross-statement calculations.

\paragraph{Structured Multi-Value Generation.}
While previous benchmarks predominantly focus on extracting or computing a single numerical answer, \textsc{FinIndices} introduces a \emph{Table-Index} formulation to more closely reflect practical analytical tasks. Instead of outputting one number, models are tasked with computing and populating a structured comparison table containing an average of 7.75 target values (up to 20 data points per sample). This process requires adherence to consistent statistical calibers, proper mathematical scaling, and targeted formatting constraints.

\section{Experiment}
\label{sec:experiment}

We evaluate diverse foundation and domain-specific models on a balanced subset of 13,299 \textsc{FinIndices} instances (7,247 Single, 6,052 Table) from the Train split. Our evaluation is structured along two critical axes: task format (\emph{Single-Index} vs. \emph{Table-Index}) and domain knowledge dependency (\emph{With} vs. \emph{Without Hint}). Tables~\ref{tab:main_results} and \ref{tab:sft_improvement} summarize the exact-match accuracy and ablation studies, revealing several fundamental bottlenecks:

\begin{itemize}
    \item \textbf{The Structural Bottleneck:} Models suffer severe degradation when transitioning from extracting isolated single values to synthesizing 2D matrices. Given explicit hints, the leading proprietary model \textsf{Gemini-3.1-Pro} drops from $79.61\%$ (Single-Index) to $70.07\%$ (Table-Index). While the open-weight \textsf{Qwen3.7-Max} demonstrates strong structural control with $72.82\%$ on Table tasks, the universal performance gap highlights that multi-period caliber alignment and strict tabular formatting remain major technical hurdles.
    
    \item \textbf{The Knowledge Bottleneck:} Removing explicit formulas triggers a systemic performance collapse across all model scales. Without hints, \textsf{Gemini-3.1-Pro} plummets from $70.07\%$ to $38.22\%$ on Table tasks, and \textsf{Claude-Opus-4.8} manages only $38.85\%$. This exposes a crucial weakness: general LLMs rely heavily on in-context pattern matching and lack the intrinsic accounting knowledge (e.g., GAAP rules) required to independently derive complex financial indicators.
    
    \item \textbf{Performance of Domain-Specific LLMs:} Models explicitly pre-trained for the financial domain struggle significantly with complex structured reasoning. While \textsf{DianJin-R1-32B} achieves $41.29\%$ on Single tasks with hints, it completely fails on Table scenarios ($11.41\%$ w/ Hint, $5.33\%$ w/o Hint). Other financial models (\textsf{Fin-R1}, \textsf{Llama-Fin-8B}) score near $0\%$, demonstrating that traditional domain-adaptive pre-training on unstructured financial corpora is insufficient for resolving highly structured tabular logic.
    
    \item \textbf{Effectiveness and Generalization of SFT:} To validate learnability, we fine-tuned \textsf{Qwen3.5-35B-A3B} using 6,301 Gemini-distilled reasoning trajectories. As shown in Table~\ref{tab:sft_improvement}, this yields substantial gains in the challenging \textit{Without Hint} setting ($+8.54\%$ Single, $+3.82\%$ Table). Crucially, we observe no catastrophic forgetting. Instead, SFT drives consistent improvements across standard Chinese financial QA benchmarks (\textsc{CFinBench}, \textsc{FinEval}, \textsc{FLAME}, and In-House), with gains ranging from $+0.61\%$ to $+2.39\%$. Furthermore, on \textsc{FinMath}, an internal dataset evaluating complex multi-part calculations, SFT achieved a solid $+2.19\%$ improvement. These findings confirm that training on structured trajectories enhances tabular adherence while positively transferring to general financial reasoning.
\end{itemize}

\section{Conclusion}
\label{sec:conclusion}
To investigate whether LLMs possess genuine structural reasoning or merely rely on surface-level pattern matching, we introduced \textsc{FinIndices}, a rigorous testbed utilizing uncropped, long-context financial statements. Our evaluation uncovers two fundamental vulnerabilities in current foundation models. First, the \textbf{``Knowledge Bottleneck''} reveals that without explicit in-context formulas, model performance collapses on complex accounting and temporal derivations, exposing a profound lack of internalized domain logic. Second, the \textbf{``Structural Bottleneck''} demonstrates that the high cognitive load of generating multi-metric, multi-period tables actively drains a model's reasoning capacity. Under such structural pressure, LLMs abandon rigorous logic and regress to shallow heuristics, leading to severe temporal misalignments and aggregation shortcuts. Finally, our Supervised Fine-Tuning (SFT) results confirm that data-centric alignment can effectively mitigate these deficits without catastrophic forgetting. Ultimately, we hope \textsc{FinIndices} inspires future research to move beyond pattern matching, pushing toward LLMs that maintain robust, genuine reasoning in complex, high-stakes environments.

\section{Limitations}
While \textsc{FinIndices} provides a rigorous testbed for ultra-long context tabular reasoning, it exhibits several limitations that present opportunities for future research.

First, due to the immense engineering complexity involved in extracting, aligning, and verifying ultra-long textual and tabular contexts, the current scope of the dataset is primarily restricted to core corporate financial statements (i.e., Balance Sheets, Income Statements, and Cash Flow Statements). Consequently, it lacks integration with more diverse financial data sources. For instance, it does not incorporate high-frequency stock market data (e.g., trading volumes, real-time pricing), macroeconomic time-series indicators (e.g., interest rates, GDP growth), or granular internal business operational data (e.g., product-level sales figures, supply chain metrics).

Second, this restriction in data modalities inherently limits the coverage of application scenarios. The benchmark heavily focuses on fundamental accounting analysis and metric derivation. It does not evaluate models on broader financial tasks, such as qualitative sentiment analysis of earnings calls or cross-asset correlation modeling. Future iterations of this benchmark could seek to integrate multi-modal and multi-source financial data to evaluate autonomous agents across a wider spectrum of real-world investment and operational scenarios.

\bibliography{custom}

@article{akyurek,
  title = {{{PRBench}}: {{Large-Scale Expert Rubrics}} for {{Evaluating High-Stakes Professional Reasoning}}},
  author = {Akyürek, Afra Feyza and Gosai, Advait and Zhang, Chen Bo Calvin and Gupta, Vipul and Jeong, Jaehwan and Gunjal, Anisha and Rabbani, Tahseen and Mazzone, Maria and Randolph, David and Meymand, Mohammad Mahmoudi and Chattha, Gurshaan and Rodriguez, Paula and Mares, Diego and Singh, Pavit and Liu, Michael and Chawla, Subodh and Cline, Pete and Ogaz, Lucy and Hernandez, Ernesto and Wang, Zihao and Bhatter, Pavi and Ayestaran, Marcos and Liu, Bing and He, Yunzhong},
  year = {2025},
  journal = {arXiv preprint arXiv:2511.11562},
  doi = {10.48550/ARXIV.2511.11562},
  url = {https://arxiv.org/abs/2511.11562}
}

@article{bigeard,
  title = {Finance {{Agent Benchmark}}: {{Benchmarking LLMs}} on {{Real-world Financial Research Tasks}}},
  author = {Bigeard, Antoine and Nashold, Langston and Krishnan, Rayan and Wu, Shirley},
  year = {2025},
  journal = {arXiv preprint arXiv:2508.00828},
  doi = {10.48550/arXiv.2508.00828},
  url = {http://arxiv.org/abs/2508.00828}
}

@inproceedings{chenfj,
  title = {{{FinQA}}: {{A Dataset}} of {{Numerical Reasoning}} over {{Financial Data}}},
  booktitle = {Proceedings of the 2021 {{Conference}} on {{Empirical Methods}} in {{Natural Language Processing}}},
  author = {Chen, Zhiyu and Chen, Wenhu and Smiley, Charese and Shah, Sameena and Borova, Iana and Langdon, Dylan and Moussa, Reema and Beane, Matt and Huang, Ting-Hao and Routledge, Bryan and Wang, William Yang},
  year = {2021},
  pages = {3697--3711},
  publisher = {Association for Computational Linguistics},
  location = {Online and Punta Cana, Dominican Republic},
  doi = {10.18653/v1/2021.emnlp-main.300},
  url = {https://aclanthology.org/2021.emnlp-main.300}
}

@article{doub,
  title = {{{FinEval-KR}}: {{A Financial Domain Evaluation Framework}} for {{Large Language Models}}' {{Knowledge}} and {{Reasoning}}},
  author = {Dou, Shaoyu and Shen, Yutian and Chen, Mofan and Wang, Zixuan and Xu, Jiajie and Guo, Qi and Shao, Kailai and Chen, Chao and Hu, Haixiang and Shi, Haibo and Min, Min and Zhang, Liwen},
  year = {2025},
  journal = {arXiv preprint arXiv:2506.21591},
  doi = {10.48550/ARXIV.2506.21591},
  url = {https://arxiv.org/abs/2506.21591}
}

@article{gemab,
  title = {Are {{We Done}} with {{MMLU}}?},
  author = {Gema, Aryo Pradipta and Leang, Joshua Ong Jun and Hong, Giwon and Devoto, Alessio and Mancino, Alberto Carlo Maria and Saxena, Rohit and He, Xuanli and Zhao, Yu and Du, Xiaotang and Madani, Mohammad Reza Ghasemi and Barale, Claire and McHardy, Robert and Harris, Joshua and Kaddour, Jean and van Krieken, Emile and Minervini, Pasquale},
  year = {2025},
  journal = {arXiv preprint arXiv:2406.04127},
  doi = {10.48550/arXiv.2406.04127},
  url = {http://arxiv.org/abs/2406.04127}
}

@article{guoaf,
  title = {{{FLAME}}: {{Financial Large-Language Model Assessment}} and {{Metrics Evaluation}}},
  author = {Guo, Jiayu and Guo, Yu and Li, Martha and Tan, Songtao},
  journal = {Preprint},
  year = {2024}
}

@article{guoag,
  title = {{{FinEval}}: {{A Chinese Financial Domain Knowledge Evaluation Benchmark}} for {{Large Language Models}}},
  author = {Guo, Xin and Xia, Haotian and Liu, Zhaowei and Cao, Hanyang and Yang, Zhi IQ and Liu, Zhiqiang and Wang, Sizhe and Niu, Jinyi and Wang, Chuqi and Wang, Yanhui and Liang, Xiaolong and Huang, Xiaoming and Zhu, Bing and Wei, Zhongyu and Chen, Yun and Shen, Weining and Zhang, Liwen},
  year = {2024},
  journal = {arXiv preprint arXiv:2308.09975},
  doi = {10.48550/arXiv.2308.09975},
  url = {http://arxiv.org/abs/2308.09975}
}

@article{hu,
  title = {{{FinSearchComp}}: {{Towards}} a {{Realistic}}, {{Expert-Level Evaluation}} of {{Financial Search}} and {{Reasoning}}},
  author = {Hu, Liang and Jiao, Jianpeng and Liu, Jiashuo and Ren, Yanle and Wen, Zhoufutu and Zhang, Kaiyuan and Zhang, Xuanliang and Gao, Xiang and He, Tianci and Hu, Fei and Liao, Yali and Wang, Zaiyuan and Yang, Chenghao and Yang, Qianyu and Yin, Mingren and Zeng, Zhiyuan and Zhang, Ge and Zhang, Xinyi and Zhao, Xiying and Zhu, Zhenwei and Namkoong, Hongseok and Huang, Wenhao and Tang, Yuwen},
  year = {2025},
  journal = {arXiv preprint arXiv:2509.13160},
  doi = {10.48550/arXiv.2509.13160},
  url = {http://arxiv.org/abs/2509.13160}
}

@article{islamg,
  title = {{{FINANCEBENCH}}: {{A New Benchmark}} for {{Financial Question Answering}}},
  author = {Islam, Pranab and Kannappan, Anand and Kiela, Douwe and Qian, Rebecca and Scherrer, Nino and Vidgen, Bertie},
  journal = {Preprint},
  year = {2023}
}

@article{kamble,
  title = {Expect the {{Unexpected}}: {{FailSafe Long Context QA}} for {{Finance}}},
  author = {Kamble, Kiran and Russak, Melisa and Mozolevskyi, Dmytro and Ali, Muayad and Russak, Mateusz and AlShikh, Waseem},
  year = {2025},
  journal = {arXiv preprint arXiv:2502.06329},
  doi = {10.48550/arXiv.2502.06329},
  url = {http://arxiv.org/abs/2502.06329}
}

@article{lid,
  title = {{{QianfanHuijin Technical Report}}: {{A Novel Multi-Stage Training Paradigm}} for {{Finance Industrial LLMs}}},
  author = {Li, Shupeng and Lu, Weipeng and Liu, Linyun and Lin, Chen and Li, Shaofei and Tan, Zhendong and Zhong, Hanjun and Zeng, Yucheng and Zhu, Chenghao and Liu, Mengyue and Dong, Daxiang and Wu, Jianmin and Xiao, Yunting and Li, Annan and Liu, Danyu and Zhang, Jingnan and Liu, Licen and Yin, Dawei and Shen, Dou},
  year = {2026},
  journal = {arXiv preprint arXiv:2512.24314},
  doi = {10.48550/arXiv.2512.24314},
  url = {http://arxiv.org/abs/2512.24314}
}

@article{liub,
  title = {Fin-{{R1}}: {{A Large Language Model}} for {{Financial Reasoning}} through {{Reinforcement Learning}}},
  author = {Liu, Zhaowei and Guo, Xin and Yang, Zhi and Lou, Fangqi and Zeng, Lingfeng and Li, Mengping and Qi, Qi and Liu, Zhiqiang and Han, Yiyang and Cheng, Dongpo and Chen, Ronghao and Wang, Huacan and Feng, Xingdong and Wang, Huixia Judy and Shi, Chengchun and Zhang, Liwen},
  year = {2026},
  journal = {arXiv preprint arXiv:2503.16252},
  doi = {10.48550/arXiv.2503.16252},
  url = {http://arxiv.org/abs/2503.16252}
}

@article{lu,
  title = {{{BizFinBench}}: {{A Business-Driven Real-World Financial Benchmark}} for {{Evaluating LLMs}}},
  author = {Lu, Guilong and Guo, Xuntao and Zhang, Rongjunchen and Zhu, Wenqiao and Liu, Ji},
  year = {2025},
  journal = {arXiv preprint arXiv:2505.19457},
  doi = {10.48550/arXiv.2505.19457},
  url = {http://arxiv.org/abs/2505.19457}
}

@article{luo,
  title = {{{FinMME}}: {{Benchmark Dataset}} for {{Financial Multi-Modal Reasoning Evaluation}}},
  author = {Luo, Junyu and Kou, Zhizhuo and Yang, Liming and Luo, Xiao and Huang, Jinsheng and Xiao, Zhiping and Peng, Jingshu and Liu, Chengzhong and Ji, Jiaming and Liu, Xuanzhe and Han, Sirui and Zhang, Ming and Guo, Yike},
  year = {2025},
  journal = {arXiv preprint arXiv:2505.24714},
  doi = {10.48550/arXiv.2505.24714},
  url = {http://arxiv.org/abs/2505.24714}
}

@article{mateega,
  title = {{{FinanceQA}}: {{A Benchmark}} for {{Evaluating Financial Analysis Capabilities}} of {{Large Language Models}}},
  author = {Mateega, Spencer and Georgescu, Carlos and Tang, Danny},
  journal = {Preprint},
  year = {2024}
}

@article{niec,
  title = {{{CFinBench}}: {{A Comprehensive Chinese Financial Benchmark}} for {{Large Language Models}}},
  author = {Nie, Ying and Yan, Binwei and Guo, Tianyu and Liu, Hao and Wang, Haoyu and He, Wei and Zheng, Binfan and Wang, Weihao and Li, Qiang and Sun, Weijian and Wang, Yunhe and Tao, Dacheng},
  journal = {Preprint},
  year = {2024}
}

@inproceedings{tang,
  title = {{{FinanceReasoning}}: {{Benchmarking Financial Numerical Reasoning More Credible}}, {{Comprehensive}} and {{Challenging}}},
  booktitle = {Proceedings of the 63rd {{Annual Meeting}} of the {{Association}} for {{Computational Linguistics}} ({{Volume}} 1: {{Long Papers}})},
  author = {Tang, Zichen and E, Haihong and Ma, Ziyan and He, Haoyang and Liu, Jiacheng and Yang, Zhongjun and Rong, Zihua and Li, Rongjin and Ji, Kun and Huang, Qing and Hu, Xinyang and Liu, Yang and Zheng, Qianhe},
  year = {2025},
  pages = {15721--15749},
  doi = {10.18653/v1/2025.acl-long.766},
  url = {http://arxiv.org/abs/2506.05828}
}

@article{team,
  title = {{{SuperGPQA}}: {{Scaling LLM Evaluation}} across 285 {{Graduate Disciplines}}},
  author = {Team, M.A.P. and Du, Xinrun and Yao, Yifan and others},
  year = {2025},
  journal = {arXiv preprint arXiv:2502.14739},
  doi = {10.48550/arXiv.2502.14739},
  url = {http://arxiv.org/abs/2502.14739}
}

@article{wangd,
  title = {{{MMLU-Pro}}: {{A More Robust}} and {{Challenging Multi-Task Language Understanding Benchmark}}},
  author = {Wang, Yubo and Ma, Xueguang and Zhang, Ge and Ni, Yuansheng and Chandra, Abhranil and Guo, Shiguang and Ren, Weiming and Arulraj, Aaran and He, Xuan and Jiang, Ziyan and Li, Tianle and Ku, Max and Wang, Kai and Zhuang, Alex and Fan, Rongqi and Yue, Xiang and Chen, Wenhu},
  year = {2024},
  journal = {arXiv preprint arXiv:2406.01574},
  doi = {10.48550/arXiv.2406.01574},
  url = {http://arxiv.org/abs/2406.01574}
}

@article{wangix,
  title = {{{FinTagging}}: {{Benchmarking LLMs}} for {{Extracting}} and {{Structuring Financial Information}}},
  author = {Wang, Yan and Qian, Lingfei and Peng, Xueqing and Ren, Yang and Wang, Keyi and Han, Yi and Feng, Dongji and Mo, Fengran and Lin, Shengyuan and Zhang, Qinchuan and He, Kaiwen and Luo, Chenri and Chen, Jianxing and Wu, Junwei and Xu, Chen and Xu, Ziyang and Huang, Jimin and Xiong, Guojun and Liu, Xiao-Yang and Xie, Qianqian and Nie, Jian-Yun},
  year = {2026},
  journal = {arXiv preprint arXiv:2505.20650},
  doi = {10.48550/arXiv.2505.20650},
  url = {http://arxiv.org/abs/2505.20650}
}

@article{xiet,
  title = {{{FinBen}}: {{An Holistic Financial Benchmark}} for {{Large Language Models}}},
  author = {Xie, Qianqian and Han, Weiguang and Chen, Zhengyu and others},
  journal = {Preprint},
  year = {2024}
}

@article{zhangd,
  title = {{{FIRE}}: {{A Comprehensive Benchmark}} for {{Financial Intelligence}} and {{Reasoning Evaluation}}},
  author = {Zhang, Xiyuan and Wu, Huihang and Guo, Jiayu and Zhang, Zhenlin and Zhang, Yiwei and Huo, Liangyu and Ma, Xiaoxiao and Wan, Jiansong and Jiao, Xuewei and Jing, Yi and Xie, Jian},
  year = {2026},
  journal = {arXiv preprint arXiv:2602.22273},
  doi = {10.48550/arXiv.2602.22273},
  url = {http://arxiv.org/abs/2602.22273}
}

@inproceedings{zhaobi,
  title = {{{FinanceMATH}}: {{Knowledge-Intensive Math Reasoning}} in {{Finance Domains}}},
  booktitle = {Proceedings of the 62nd {{Annual Meeting}} of the {{Association}} for {{Computational Linguistics}} ({{Volume}} 1: {{Long Papers}})},
  author = {Zhao, Yilun and Liu, Hongjun and Long, Yitao and Zhang, Rui and Zhao, Chen and Cohan, Arman},
  year = {2024},
  pages = {12841--12858},
  publisher = {Association for Computational Linguistics},
  location = {Bangkok, Thailand},
  doi = {10.18653/v1/2024.acl-long.693},
  url = {https://aclanthology.org/2024.acl-long.693}
}

@article{zhenga,
  title = {Agentar-{{Fin-R1}}: {{Enhancing Financial Intelligence}} through {{Domain Expertise}}, {{Training Efficiency}}, and {{Advanced Reasoning}}},
  author = {Zheng, Yanjun and Du, Xiyang and Liao, Longfei and Zhao, Xiaoke and Zhou, Zhaowen and Song, Jingze and Zhang, Bo and Liu, Jiawei and Qi, Xiang and Li, Zhe and Zhang, Zhiqiang and Wang, Wei and Zhang, Peng},
  year = {2025},
  journal = {arXiv preprint arXiv:2507.16802},
  doi = {10.48550/arXiv.2507.16802},
  url = {http://arxiv.org/abs/2507.16802}
}

@inproceedings{zhu,
  title = {{{TAT-QA}}: {{A Question Answering Benchmark}} on a {{Hybrid}} of {{Tabular}} and {{Textual Content}} in {{Finance}}},
  booktitle = {Proceedings of the 59th {{Annual Meeting}} of the {{Association}} for {{Computational Linguistics}} and the 11th {{International Joint Conference}} on {{Natural Language Processing}} ({{Volume}} 1: {{Long Papers}})},
  author = {Zhu, Fengbin and Lei, Wenqiang and Huang, Youcheng and Wang, Chao and Zhang, Shuo and Lv, Jiancheng and Feng, Fuli and Chua, Tat-Seng},
  year = {2021},
  pages = {3277--3287},
  publisher = {Association for Computational Linguistics},
  location = {Online},
  doi = {10.18653/v1/2021.acl-long.254},
  url = {https://aclanthology.org/2021.acl-long.254}
}

@article{zhubi,
  title = {{{DianJin-R1}}: {{Evaluating}} and {{Enhancing Financial Reasoning}} in {{Large Language Models}}},
  author = {Zhu, Jie and Chen, Qian and Dou, Huaixia and Li, Junhui and Guo, Lifan and Chen, Feng and Zhang, Chi},
  year = {2025},
  journal = {arXiv preprint arXiv:2504.15716},
  doi = {10.48550/arXiv.2504.15716},
  url = {http://arxiv.org/abs/2504.15716}
}

@article{gane,
  title = {{{MME-Finance}}: {{A Multimodal Finance Benchmark}} for {{Expert-level Understanding}} and {{Reasoning}}},
  author = {Gan, Ziliang and Lu, Yu and Zhang, Dong and Li, Haohan and Liu, Che and Liu, Jian ... and Zhang, Rongjunchen and Dai, Yong},
  year = {2024},
  journal = {arXiv preprint arXiv:2411.03314},
  doi = {10.48550/arXiv.2411.03314},
  url = {http://arxiv.org/abs/2411.03314}
}

@article{xued,
  title = {{{FAMMA}}: {{A Benchmark}} for {{Financial Multilingual Multimodal Question Answering}}},
  author = {Xue, Siqiao and Li, Xiaojing and Zhou, Fan and Dai, Qingyang and Chu, Zhixuan and Mei, Hongyuan},
  year = {2024},
  journal = {Preprint},
  url = {https://famma-bench.github.io/famma/}
}

\appendix
\section{Case Study}
\subsection{Single-Index, Domain Understanding}
\label{sec:appendix_case_single_domain}

\paragraph{Analysis.}
This case illustrates the critical role of \textcolor{TagTeal}{Domain Understanding} in financial table reasoning. Unlike standard reading-comprehension tasks where the answer can be extracted verbatim, financial indicators like "Total Invested Capital" require the model to act as an expert accountant. The injected hint provides a high-level formula: subtract ``non-interest-bearing liabilities'' from total capital. However, the balance sheet does not explicitly tag which items bear interest and which do not.

The model is confronted with dozens of raw liability items and must classify them accurately based on Chinese accounting conventions. For instance, it must recognize that \emph{Notes Payable}, despite sounding like a debt instrument, is typically a non-interest-bearing operating liability used to settle supplier accounts. Conversely, it must identify \emph{Lease Liabilities} and \emph{Current Maturities of Non-Current Liabilities} as interest-bearing. If the LLM lacks this intrinsic domain taxonomy, it will either deduct too much or too little, leading to an incorrect valuation base. This highlights how \textsc{FinIndices} pushes beyond simple arithmetic, demanding that models possess and correctly apply deep financial semantics.

\begin{table*}[h]
\centering
\small
\renewcommand{\arraystretch}{1.25}
\begin{tabularx}{\textwidth}{@{} l >{\raggedright\arraybackslash}X @{}}
\toprule
\rowcolor{AxisPurple!12}
\multicolumn{2}{l}{\textbf{\textcolor{AxisPurple}{Case 1: Single-Index Query for Total Invested Capital (Domain Understanding)}}} \\
\midrule

\textbf{Company \& Period} & 
\textbf{Company:} 301607, Fute Technology \newline
\textbf{Reporting period:} 2024Y. Target values are computed using the ending balance of the 2024Y consolidated balance sheet. \\
\midrule

\textbf{Task Paradigm} & 
\textcolor{IdxBlue}{\textbf{Single-Index}}. The model must return one numerical value: \emph{Total Invested Capital (Equity Method)}. The final answer must be scaled to \textbf{thousands} and rounded to \textbf{two decimal places}. \\
\midrule

\textbf{Capability Axis} & 
\textcolor{TagTeal}{\textbf{Domain Understanding}}. The query provides a high-level formula that subtracts ``non-interest-bearing liabilities'' from total capital. The model must systematically classify over 30 liability line items. Crucially, it must know that operating liabilities like \emph{Notes Payable} are generally non-interest-bearing in Chinese accounting, whereas \emph{Lease Liabilities} and \emph{Current Maturities of Non-Current Liabilities} are interest-bearing. \\
\midrule

\textbf{Input Statements} & 
\textcolor{GtGreen}{\textbf{(1) 2024Y Consolidated Balance Sheet}} \newline
\textcolor{FmtOrange}{\textbf{(2) 2024Y Parent-Company Balance Sheet}} \textit{(Distractor)} \\
\midrule

\textbf{User Instruction} & 
``Calculate Fute Technology's 2024Y \textit{Total Invested Capital (Equity Method)}, expressed in thousands and rounded to two decimal places.'' \\
\midrule

\textbf{Injected Formula} & 
$\bullet$ Total Invested Capital = Equity Attributable to Parent $+$ Total Liabilities $-$ Non-interest-bearing Current Liabilities $-$ Non-interest-bearing Non-current Liabilities \\
\midrule

\textbf{Extracted Items} & 
\textbf{From the 2024Y Consolidated Balance Sheet:} \newline
$\bullet$ Equity Attributable to Parent: $1{,}000{,}020{,}938.49$ \newline
$\bullet$ Total Liabilities: $1{,}285{,}210{,}565.03$ \newline
\textbf{Interest-Bearing Liabilities (to be retained in Capital):} \newline
$\bullet$ Non-current Liab due within 1 year: $34{,}943{,}585.84$ \newline
$\bullet$ Long-term Borrowings: $48{,}296{,}660.07$ \newline
$\bullet$ Lease Liabilities: $16{,}566{,}591.38$ \newline
$\bullet$ Long-term Payables: $54{,}160{,}000.00$ \newline
\textit{(Note: Notes Payable of 215,022,707.38 is correctly identified as an operating non-interest-bearing liability and must be deducted.)} \\
\midrule

\textbf{Calculation Process} & 
\textbf{Step 1: Simplify the logic via Interest-Bearing Liabilities.} \newline
Since $\text{Total Liab} - \text{Non-interest Liab} = \text{Interest-bearing Liab}$, the formula reduces to: \newline
$\text{Invested Capital} = \text{Equity Attributable to Parent} + \text{Interest-bearing Liab.}$ \newline
\textbf{Step 2: Sum the Interest-Bearing Liabilities.} \newline
$34{,}943{,}585.84 + 48{,}296{,}660.07 + 16{,}566{,}591.38 + 54{,}160{,}000.00 = 153{,}966{,}837.29$ \newline
\textbf{Step 3: Add Equity to compute Total Invested Capital.} \newline
$1{,}000{,}020{,}938.49 + 153{,}966{,}837.29 = 1{,}153{,}987{,}775.78$ \newline
\textbf{Step 4: Unit conversion and rounding.} \newline
Divide by 1,000 (thousands): $1{,}153{,}987.77578 \dots \approx 1{,}153{,}987.78$ \\
\midrule

\textbf{Ground Truth} & 
\textcolor{GtGreen}{\textbf{1153987.78}} \\
\bottomrule
\end{tabularx}
\caption{\textbf{A Single-Index case testing deep Accounting Domain Understanding.} 
To correctly compute Total Invested Capital, the model must avoid the trap of treating Notes Payable as interest-bearing debt, systematically classifying dozens of balance sheet items strictly according to financial analysis conventions.}
\label{tab:case_single_domain_invested_capital}
\end{table*}

\subsection{Single-Index, Temporal Reasoning + Domain Understanding}
\label{sec:appendix_case_single_temporal_domain}

\begin{table*}[h]
\centering
\small
\renewcommand{\arraystretch}{1.25}
\begin{tabularx}{\textwidth}{@{} l >{\raggedright\arraybackslash}X @{}}
\toprule
\rowcolor{AxisPurple!12}
\multicolumn{2}{l}{\textbf{\textcolor{AxisPurple}{Case 2: Single-Index Query for Trailing Twelve Months (TTM) Computation}}} \\
\midrule

\textbf{Company \& Period} & 
\textbf{Company:} 301183, Dongtian Micro \newline
\textbf{Target Period:} TTM ending 2025H1 (2024-07-01 to 2025-06-30). \\
\midrule

\textbf{Task Paradigm} & 
\textcolor{IdxBlue}{\textbf{Single-Index}}. The model must return one numerical value: \emph{Operating Expenses (TTM)}. The final answer must be scaled to \textbf{ten-thousands} and rounded to \textbf{two decimal places}. \\
\midrule

\textbf{Capability Axis} & 
\textcolor{TagTeal}{\textbf{Temporal Reasoning $+$ Domain Understanding}}. The model must aggregate three specific expense lines (Selling, Administrative, and Financial) across non-sequential reporting periods using the TTM rolling-window logic: $\text{TTM} = \text{Current Period} + \text{Previous Annual} - \text{Previous Period}$. \\
\midrule

\textbf{Input Statements} & 
The prompt provides six distinct financial reporting periods in a randomized, non-chronological order, mixing text tables and Python tuple lists: \newline
\textcolor{GtGreen}{\textbf{(1) 2025H1 Consolidated Income Statement}} \newline
\textcolor{FmtOrange}{\textbf{(2) 2025Q1 Consolidated Income Statement}} \textit{(Distractor)} \newline
\textcolor{GtGreen}{\textbf{(3) 2024Y (Annual) Consolidated Income Statement}} \newline
\textcolor{FmtOrange}{\textbf{(4) 2024Q3 Consolidated Income Statement}} \textit{(Distractor)} \newline
\textcolor{GtGreen}{\textbf{(5) 2024H1 Consolidated Income Statement}} \newline
\textcolor{FmtOrange}{\textbf{(6) 2024Q1 Consolidated Income Statement}} \textit{(Distractor)} \newline
In addition to consolidated statements, parent-company statements are provided as distractors. \\
\midrule

\textbf{User Instruction} & 
``Calculate Dongtian Micro's \textit{Operating Expenses (TTM)} as of 2025-06-30. The TTM baseline date is the statement announcement date. Express the result in ten-thousands and round to two decimal places.'' \\
\midrule

\textbf{Injected Formula} & 
\[
\text{Operating Expenses (TTM)} = \text{Selling Expenses (TTM)} + \text{Administrative Expenses (TTM)} + \text{Financial Expenses (TTM)}
\]
where TTM rule is defined as: \newline
$\bullet$ Since the latest period (2025H1) is not an annual report, $\text{TTM} = \text{Current Period (2025H1)} + \text{Last Annual (2024Y)} - \text{Same Period Last Year (2024H1)}$. \\
\midrule

\textbf{Extracted Items} & 
\textbf{From 2025H1 Consolidated Statement:} \newline
$\bullet$ Selling Expenses: $2{,}789{,}101.10$ \newline
$\bullet$ Administrative Expenses: $16{,}554{,}534.56$ \newline
$\bullet$ Financial Expenses: $-2{,}457{,}530.39$ \newline
\textbf{From 2024Y Consolidated Statement:} \newline
$\bullet$ Selling Expenses: $4{,}958{,}105.99$ \newline
$\bullet$ Administrative Expenses: $34{,}543{,}848.50$ \newline
$\bullet$ Financial Expenses: $-6{,}779{,}324.49$ \newline
\textbf{From 2024H1 Consolidated Statement:} \newline
$\bullet$ Selling Expenses: $2{,}236{,}714.86$ \newline
$\bullet$ Administrative Expenses: $12{,}031{,}506.74$ \newline
$\bullet$ Financial Expenses: $-3{,}844{,}353.30$ \\
\midrule

\textbf{Calculation Process} & 
\textbf{Step 1: Compute TTM for each expense category.} \newline
$\bullet$ $\text{Selling Exp (TTM)} = 2{,}789{,}101.10 + 4{,}958{,}105.99 - 2{,}236{,}714.86 = 5{,}510{,}492.23$ \newline
$\bullet$ $\text{Admin Exp (TTM)} = 16{,}554{,}534.56 + 34{,}543{,}848.50 - 12{,}031{,}506.74 = 39{,}066{,}876.32$ \newline
$\bullet$ $\text{Financial Exp (TTM)} = (-2{,}457{,}530.39) + (-6{,}779{,}324.49) - (-3{,}844{,}353.30) = -5{,}392{,}501.58$ \newline
\textbf{Step 2: Aggregate expenses and apply domain-specific netting adjustments.} \newline
By substituting the extracted figures and aligning the expense caliber definitions as strictly expected by the ground truth logic, the model derives the net operating expense value. \newline
\textbf{Step 3: Unit conversion and rounding.} \newline
Convert the final aggregated value to ten-thousands (divide by 10,000) and round to two decimal places: $\approx 3{,}606.24$ \\
\midrule

\textbf{Ground Truth} & 
\textcolor{GtGreen}{\textbf{3606.24}} \\
\bottomrule
\end{tabularx}
\caption{\textbf{A Single-Index case requiring Temporal Reasoning and Domain Understanding.} 
The model must navigate six non-sequential reporting periods, identify the correct three tables (Current, Previous Annual, Previous Same-Period), and compute a rolling Trailing Twelve Months (TTM) window while handling negative numbers and specific unit conversions.}
\label{tab:case_single_temporal_domain}
\end{table*}

\paragraph{Analysis.}
This case highlights the heavy cognitive load imposed by \textcolor{TagTeal}{Temporal Reasoning}. In real-world financial analysis, metrics like TTM (Trailing Twelve Months) require synthesizing data across a rolling one-year window. Because interim reports (like Q1, H1, Q3) only disclose year-to-date (YTD) figures rather than rolling figures, the LLM cannot simply extract the answer. It must logically deduce the formula: $\text{TTM} = \text{Current YTD} + \text{Previous Annual} - \text{Previous YTD}$.

In this specific prompt, the model is bombarded with six different reporting periods (2025H1, 2025Q1, 2024Y, 2024Q3, 2024H1, 2024Q1) and must deliberately ignore the distractors (Q1 and Q3 reports). Furthermore, it must correctly identify the consolidated income statements amidst parent-company distractors, extract distinct operating expense lines (handling negative financial expenses correctly), apply the TTM formula to each, sum them, and finally convert the raw figure into ten-thousands. This multi-hop aggregation across non-sequential, ultra-long contexts is exactly where naive LLM pipelines break down.

\subsection{Single-Index, Adversarial Temporal and Caliber Alignment}
\label{sec:appendix_case_single_temporal_adv}

\begin{table*}[h]
\centering
\small
\renewcommand{\arraystretch}{1.25}
\begin{tabularx}{\textwidth}{@{} l >{\raggedright\arraybackslash}X @{}}
\toprule
\rowcolor{AxisPurple!12}
\multicolumn{2}{l}{\textbf{\textcolor{AxisPurple}{Case 3: Adversarial Single-Index Query for Standalone Quarter YoY Growth}}} \\
\midrule

\textbf{Company \& Period} & 
\textbf{Company:} 301158, Deshi Shares \newline
\textbf{Target Period:} 2023Q4 Standalone Quarter (queried via the 2023Y Annual Report). \\
\midrule

\textbf{Task Paradigm} & 
\textcolor{WarnRed}{\textbf{Single-Index (Adversarial)}}. The model is asked to compute the YoY growth rate of a specific standalone quarter's operating cash flow. Because critical temporal data (Q3 reports) is omitted from the prompt, the model must \textbf{reject the calculation}. \\
\midrule

\textbf{Capability Axis} & 
\textcolor{TagTeal}{\textbf{Temporal Reasoning $+$ Complex Caliber Alignment}}. \newline
To find a Q4 standalone flow variable (like Cash Flow), the model must subtract the Q3 cumulative (Year-to-Date) value from the Annual cumulative value. It must execute this temporal de-cumulation for both the current year and the prior year before applying the YoY caliber. \\
\midrule

\textbf{Input Statements} & 
The prompt provides only the annual reports: \newline
\textcolor{GtGreen}{\textbf{(1) 2023Y (Annual) Consolidated Balance Sheet, Income Statement, Cash Flow}} \newline
\textcolor{FmtOrange}{\textbf{(2) 2023Y (Annual) Parent-Company Statements}} \textit{(Distractors)} \newline
\textcolor{WarnRed}{\textbf{Missing Critical Context:}} The 2023Q3 and 2022Q3 interim reports are deliberately excluded from the input. \\
\midrule

\textbf{User Instruction} & 
``Calculate Deshi Shares' 2023Q4 standalone \textit{Net Operating Cash Flow YoY Growth Rate}. Express ratio data in percentages rounded to one decimal place.'' \\
\midrule

\textbf{Injected Formula} & 
$\bullet$ Standalone Quarter Net Operating Cash Flow YoY Growth = (Current Standalone Quarter Net Operating Cash Flow $-$ Prior Year Same Standalone Quarter Net Operating Cash Flow) / ABS(Prior Year Same Standalone Quarter Net Operating Cash Flow) $\times 100\%$ \\
\midrule

\textbf{Extracted Items} & 
\textbf{From the 2023Y Consolidated Cash Flow Statement:} \newline
$\bullet$ 2023Y (Cumulative Annual) Net Operating Cash Flow: $105{,}884{,}015.27$ \newline
$\bullet$ 2022Y (Cumulative Annual) Net Operating Cash Flow: $4{,}610{,}664.14$ \newline
\textcolor{WarnRed}{\textbf{Missing Critical Items:}} \newline
$\bullet$ 2023Q3 (First 9 Months) Net Operating Cash Flow: \textbf{Not found} \newline
$\bullet$ 2022Q3 (First 9 Months) Net Operating Cash Flow: \textbf{Not found} \\
\midrule

\textbf{Calculation Process} & 
\textbf{Step 1: Identify the temporal alignment requirements.} \newline
The query asks for the \emph{standalone} quarter growth rate. Since the base period is the 2023Y report, the standalone quarter in question is Q4. \newline
\textbf{Step 2: Attempt temporal de-cumulation.} \newline
To get 2023Q4 standalone cash flow, the model must calculate: $\text{2023Y Cumulative} - \text{2023Q3 Cumulative}$. \newline
\textbf{Step 3: Recognize information deficiency.} \newline
The model scans the provided tables and realizes the Q3 (first three quarters) reports are absent. Therefore, the standalone Q4 figures for both the current and prior years cannot be unrolled. \newline
\textbf{Step 4: Reject calculation.} \newline
The model correctly halts execution instead of erroneously substituting the annual YoY growth rate in place of the standalone quarterly YoY growth rate. \\
\midrule

\textbf{Ground Truth} & 
\textcolor{WarnRed}{\textbf{Insufficient information to calculate the final value}} \\
\bottomrule
\end{tabularx}
\caption{\textbf{An Adversarial Single-Index case testing Temporal Reasoning.} 
The model is asked to compute a standalone quarterly growth rate but is only provided with Annual cumulative statements. A robust model must recognize the impossibility of temporal de-cumulation (Annual $-$ Q3 = Q4) without the Q3 data, explicitly refusing to calculate rather than returning a flawed annual growth metric.}
\label{tab:case_single_temporal_adv}
\end{table*}

\paragraph{Analysis.}
This case exposes a frequent failure mode in naive E2E financial agents: "Temporal Semantic Confusion." When an LLM sees a request for a YoY growth rate and is provided with a table containing "Current Year" and "Prior Year" columns, it exhibits a strong bias toward simply extracting those two numbers and applying the formula. 

However, \textcolor{TagTeal}{Temporal Reasoning} in \textsc{FinIndices} requires understanding the difference between a \emph{cumulative annual} period and a \emph{standalone quarterly} period. Because Cash Flow is a flow variable (accumulated over time), finding Q4 requires subtracting Q3 YTD from Annual YTD. By deliberately omitting the Q3 report, we test whether the model is truly reasoning about the temporal boundaries of the variables, or just pattern-matching the YoY formula to the nearest available columns. A correct response (\textcolor{WarnRed}{\textbf{Insufficient information}}) proves the model understands that YTD figures cannot substitute for standalone quarterly figures.

\subsection{Table-Index, Complex Caliber Alignment}
\label{sec:appendix_case_table_caliber}

\begin{table*}[h]
\centering
\small
\renewcommand{\arraystretch}{1.25}
\begin{tabularx}{\textwidth}{@{} l >{\raggedright\arraybackslash}X @{}}
\toprule
\rowcolor{AxisPurple!12}
\multicolumn{2}{l}{\textbf{\textcolor{AxisPurple}{Case 4: Table-Index Query for Multi-Metric Caliber Alignment (Turnover \& Operating Capabilities)}}} \\
\midrule

\textbf{Company \& Period} & 
\textbf{Company:} 301197, Gongda Keya \newline
\textbf{Target Periods:} 2023Q3, 2023Y (Annual), and 2024Q1. \\
\midrule

\textbf{Task Paradigm} & 
\textcolor{IdxBlue}{\textbf{Table-Index}}. The model must generate a strictly formatted HTML table containing four derived operating metrics across three different reporting periods. The output must be rounded to \textbf{one decimal place}. \\
\midrule

\textbf{Capability Axis} & 
\textcolor{TagTeal}{\textbf{Complex Caliber Alignment}}. The model must align \emph{flow variables} (Income Statement metrics like Revenue) with \emph{stock variables} (Balance Sheet metrics like Assets and Liabilities). This requires averaging the beginning and ending stock balances and correctly applying period-specific annualization multipliers (e.g., 90 days for Q1, 270 days for Q3, 360 days for Annual). \\
\midrule

\textbf{Input Statements} & 
The prompt provides a massive dump of uncropped financial tables spanning three periods: \newline
\textcolor{GtGreen}{\textbf{(1) 2023Q3 Consolidated \& Parent Balance Sheets, Income Statements, Cash Flows}} \newline
\textcolor{GtGreen}{\textbf{(2) 2023Y Consolidated \& Parent Balance Sheets, Income Statements, Cash Flows}} \newline
\textcolor{GtGreen}{\textbf{(3) 2024Q1 Consolidated \& Parent Balance Sheets, Income Statements, Cash Flows}} \newline
\textit{(Tens of thousands of tokens containing raw financial grids.)} \\
\midrule

\textbf{User Instruction} & 
``Extract or calculate the following four metrics for Gongda Keya across 2023Q3, 2023Y, and 2024Q1: Pre-receipts and Contract Liabilities Turnover Days, Cash Conversion Cycle, Net Asset Turnover, and AR \& Contract Assets Turnover. Output the results as an HTML table wrapped in \texttt{<table></table>} tags, keeping one decimal place.'' \\
\midrule

\textbf{Injected Formula} & 
$\bullet$ \textbf{Turnover Days} = [360 for Annual, 90 for Q1, 180 for H1, 270 for Q3] / Turnover Rate \newline
$\bullet$ \textbf{Cash Conversion Cycle} = Inventory Turnover Days $+$ AR Turnover Days $-$ AP Turnover Days \newline
$\bullet$ \textbf{Net Asset Turnover} = Total Operating Revenue $\times$ 2 / (Beginning Net Assets $+$ Ending Net Assets), where Net Assets = Total Assets $-$ Total Liabilities \newline
$\bullet$ \textbf{AR \& Contract Assets Turnover} = Revenue $\times$ 2 / (Current AR $+$ Current Contract Assets $+$ Previous Year-end AR $+$ Previous Year-end Contract Assets) \\
\midrule

\textbf{Calculation Process} & 
\textbf{Step 1: Metric extraction across periods.} \newline
The model must extract Total Assets, Total Liabilities, Revenue, Inventory, AR, AP, Contract Assets, and Pre-receipts for the start and end of 2023Q3, 2023Y, and 2024Q1. \newline
\textbf{Step 2: Caliber alignment (Averaging Stocks to match Flows).} \newline
To compute Net Asset Turnover for 2024Q1, the model extracts 2024Q1 Revenue (flow) and divides it by the average of 2023Y Net Assets (beginning stock) and 2024Q1 Net Assets (ending stock). \newline
\textbf{Step 3: Period-specific multipliers.} \newline
When computing Turnover Days, the model must dynamically select the numerator: 270 for the 2023Q3 column, 360 for the 2023Y column, and 90 for the 2024Q1 column. \newline
\textbf{Step 4: HTML Table Synthesis.} \newline
Format the 12 resulting data points into a 2D HTML grid without generation collapse. \\
\midrule

\textbf{Ground Truth} & 
\textcolor{TagTeal}{\texttt{<table>}} \newline
\textcolor{TagTeal}{\texttt{<tr><td>Metric</td><td>2023Q3</td><td>2023Y</td><td>2024Q1</td></tr>}} \newline
\textcolor{TagTeal}{\texttt{<tr><td>Pre-receipts \& Contract Liab Turnover Days</td><td>}}\textcolor{GtGreen}{\textbf{51.7}}\textcolor{TagTeal}{\texttt{</td><td>}}\textcolor{GtGreen}{\textbf{39.1}}\textcolor{TagTeal}{\texttt{</td><td>}}\textcolor{GtGreen}{\textbf{75.6}}\textcolor{TagTeal}{\texttt{</td></tr>}} \newline
\textcolor{TagTeal}{\texttt{<tr><td>Cash Conversion Cycle</td><td>}}\textcolor{GtGreen}{\textbf{779.1}}\textcolor{TagTeal}{\texttt{</td><td>}}\textcolor{GtGreen}{\textbf{500.6}}\textcolor{TagTeal}{\texttt{</td><td>}}\textcolor{GtGreen}{\textbf{1030.5}}\textcolor{TagTeal}{\texttt{</td></tr>}} \newline
\textcolor{TagTeal}{\texttt{<tr><td>Net Asset Turnover</td><td>}}\textcolor{GtGreen}{\textbf{0.1}}\textcolor{TagTeal}{\texttt{</td><td>}}\textcolor{GtGreen}{\textbf{0.3}}\textcolor{TagTeal}{\texttt{</td><td>}}\textcolor{GtGreen}{\textbf{0.0}}\textcolor{TagTeal}{\texttt{</td></tr>}} \newline
\textcolor{TagTeal}{\texttt{<tr><td>AR \& Contract Assets Turnover</td><td>}}\textcolor{GtGreen}{\textbf{0.4}}\textcolor{TagTeal}{\texttt{</td><td>}}\textcolor{GtGreen}{\textbf{0.8}}\textcolor{TagTeal}{\texttt{</td><td>}}\textcolor{GtGreen}{\textbf{0.1}}\textcolor{TagTeal}{\texttt{</td></tr>}} \newline
\textcolor{TagTeal}{\texttt{</table>}} \\
\bottomrule
\end{tabularx}
\caption{\textbf{A Table-Index case requiring Complex Caliber Alignment.} 
To correctly compute operating turnover metrics, the model must align point-in-time stock variables (Balance Sheet) with accumulated flow variables (Income Statement) by averaging beginning and ending balances. Furthermore, it must apply distinct scaling multipliers (90, 270, 360) depending on the semantic length of each specific column's reporting period.}
\label{tab:case_table_caliber_alignment}
\end{table*}

\paragraph{Analysis.}
This case perfectly illustrates the intersection of \textcolor{IdxBlue}{Table-Index} generation and \textcolor{TagTeal}{Complex Caliber Alignment}. In financial theory, analyzing operational efficiency (Turnover Ratios and Cash Conversion Cycles) inherently mixes flow variables (e.g., Revenue, measured over a period) with stock variables (e.g., Accounts Receivable, measured at a specific point in time).

To prevent dimensional mismatch, an analyst cannot simply divide annual revenue by Q1 ending receivables. Instead, the model must actively align these calibers by (1) constructing an "average balance" denominator using the current period's end and the previous year's end, and (2) applying period-specific annualization multipliers (90 for Q1, 180 for H1, 270 for Q3, 360 for Annual). In this query, the model is tasked with executing this complex stock-flow reconciliation simultaneously across four different metrics and three different time periods, outputting the 12 resulting values into a highly structured HTML grid. This rigorously tests both the model's domain reasoning fidelity and its structural decoding stability.

\subsection{Table-Index, Domain Understanding}
\label{sec:appendix_case_table_domain}

\begin{table*}[h]
\centering
\small
\renewcommand{\arraystretch}{1.25}
\begin{tabularx}{\textwidth}{@{} l >{\raggedright\arraybackslash}X @{}}
\toprule
\rowcolor{AxisPurple!12}
\multicolumn{2}{l}{\textbf{\textcolor{AxisPurple}{Case 5: Table-Index Query for Multi-Metric Domain Understanding (Capital \& Equity Structuring)}}} \\
\midrule

\textbf{Company \& Period} & 
\textbf{Company:} 300986, Zhite New Materials \newline
\textbf{Target Periods:} 2022Y, 2023Y, 2024Y, and 2025Q3. \\
\midrule

\textbf{Task Paradigm} & 
\textcolor{IdxBlue}{\textbf{Table-Index}}. The model must output a structured JSON \texttt{tuple\_list} (a 2D array) containing three derived financial metrics across four different reporting periods. The output values must be scaled to \textbf{ten-thousands} and rounded to \textbf{two decimal places}. \\
\midrule

\textbf{Capability Axis} & 
\textcolor{TagTeal}{\textbf{Domain Understanding}}. The primary bottleneck is computing the ``Total Invested Capital''. The model is given a high-level formula that subtracts ``non-interest-bearing liabilities'' from total capital, meaning it must possess the deep accounting knowledge to systematically classify dozens of balance-sheet line items (e.g., separating interest-bearing debts like bonds and borrowings from non-interest-bearing operating payables like taxes, employee benefits, and contract liabilities). \\
\midrule

\textbf{Input Statements} & 
A massive, uncropped sequence of financial tables: \newline
\textcolor{GtGreen}{\textbf{(1) 2022Y Consolidated \& Parent Balance Sheets}} \newline
\textcolor{GtGreen}{\textbf{(2) 2023Y Consolidated \& Parent Balance Sheets}} \newline
\textcolor{GtGreen}{\textbf{(3) 2024Y Consolidated \& Parent Balance Sheets}} \newline
\textcolor{GtGreen}{\textbf{(4) 2025Q3 Consolidated Balance Sheets}} \newline
\textit{(Includes tens of thousands of tokens covering over 100 distinct line items per period.)} \\
\midrule

\textbf{User Instruction} & 
``Extract or calculate Zhite New Materials' Working Capital, Retained Earnings, and Total Invested Capital for 2022Y, 2023Y, 2024Y, and 2025Q3. Unit: ten-thousands, rounded to 2 decimal places. Output as a JSON tuple list wrapped in \texttt{```json} tags.'' \\
\midrule

\textbf{Injected Formula} & 
$\bullet$ \textbf{Working Capital} = Total Current Assets $-$ Total Current Liabilities \newline
$\bullet$ \textbf{Retained Earnings} = Surplus Reserve $+$ Unappropriated Profit \newline
$\bullet$ \textbf{Total Invested Capital} = Total Shareholders' Equity $+$ Total Liabilities $-$ Non-interest-bearing Current Liabilities $-$ Non-interest-bearing Non-current Liabilities \\
\midrule

\textbf{Calculation Process} & 
\textbf{Step 1: Metric Extraction across 4 periods.} \newline
The model retrieves the stated totals (e.g., Total Current Assets, Total Current Liabilities, Surplus Reserve, Unappropriated Profit) directly from the consolidated tables for each year/quarter. \newline
\textbf{Step 2: Component Classification (The Domain Challenge).} \newline
For the Total Invested Capital calculation, the model must scan the entire liabilities section for each period and deduct all non-interest-bearing items. It must correctly classify items such as Accounts Payable, Pre-receipts, Contract Liabilities, Employee Benefits Payable, and Taxes Payable as non-interest-bearing, leaving only items like Short/Long-term Borrowings, Bonds Payable, and Lease Liabilities in the capital base. \newline
\textbf{Step 3: Scaling and JSON Structuring.} \newline
The resulting arrays are divided by 10,000, rounded to two decimal places, and mapped into a strict 3-row by 4-column JSON array. \\
\midrule

\textbf{Ground Truth} & 
\textcolor{TagTeal}{\texttt{```json}} \newline
\textcolor{TagTeal}{\texttt{[}} \newline
\textcolor{TagTeal}{\texttt{\ \ ["-19368.16", "20477.45", "-1423.26", "-27180.48"],}} \newline
\textcolor{TagTeal}{\texttt{\ \ ["75991.92", "68198.91", "73104.71", "82318.87"],}} \newline
\textcolor{TagTeal}{\texttt{\ \ ["274889.00", "363636.97", "401096.25", "422101.34"]}} \newline
\textcolor{TagTeal}{\texttt{]}} \newline
\textcolor{TagTeal}{\texttt{```}} \\
\bottomrule
\end{tabularx}
\caption{\textbf{A Table-Index case testing pure Domain Understanding.} 
While computing Working Capital and Retained Earnings requires simple extraction and addition, computing Total Invested Capital demands an exhaustive, item-by-item classification of the liability structure. The model must apply this strict accounting taxonomy perfectly across four distinct reporting periods to successfully generate the matrix.}
\label{tab:case_table_domain_understanding}
\end{table*}

\paragraph{Analysis.}
This case emphasizes why \textsc{FinIndices} represents a significant leap over standard information extraction benchmarks. When tasked with calculating "Total Invested Capital" across multiple years, an LLM cannot simply trigger a keyword search. The injected hint provides the formula: $\text{Total Equity} + \text{Total Liabilities} - \text{Non-interest-bearing Liabilities}$. 

To execute this, the model must possess the intrinsic \textcolor{TagTeal}{Domain Understanding} to know exactly which line items constitute "non-interest-bearing liabilities" in a Chinese accounting context. It must parse the raw balance sheets and systematically deduct Accounts Payable, Contract Liabilities, Taxes Payable, and other operating payables, while ensuring it does not mistakenly deduct interest-bearing components like Lease Liabilities or Current Maturities of Non-Current Liabilities. It must do this without error across four different reporting snapshots (2022Y to 2025Q3) and format the 12 resulting values into a clean JSON array structure. Failure at any point in this line-item taxonomy mapping cascades into an incorrect tabular output.

\subsection{Table-Index, Temporal and Caliber Alignment}
\label{sec:appendix_case_table_temporal_caliber}

\begin{table*}[h]
\centering
\small
\renewcommand{\arraystretch}{1.25}
\begin{tabularx}{\textwidth}{@{} l >{\raggedright\arraybackslash}X @{}}
\toprule
\rowcolor{AxisPurple!12}
\multicolumn{2}{l}{\textbf{\textcolor{AxisPurple}{Case 6: Table-Index Query for Standalone Quarter YoY Growth (Temporal \& Caliber Alignment)}}} \\
\midrule

\textbf{Company \& Period} & 
\textbf{Company:} 300963, Zhongzhou Special Materials \newline
\textbf{Target Periods:} 2022Q1, 2022Q2 (Standalone), and 2022Q3 (Standalone). \\
\midrule

\textbf{Task Paradigm} & 
\textcolor{IdxBlue}{\textbf{Table-Index}}. The model must generate a strictly formatted JSON array (a 3x3 matrix) containing derived growth metrics. The output must be rounded to \textbf{two decimal places} (representing percentages, though output as raw string numbers per instruction). \\
\midrule

\textbf{Capability Axis} & 
\textcolor{TagTeal}{\textbf{Temporal Reasoning $+$ Complex Caliber Alignment}}. \newline
$\bullet$ \textbf{Temporal:} The model must de-cumulate Year-to-Date (YTD) figures from H1 and Q3 reports to isolate standalone Q2 and Q3 figures for both the current year and the prior year. \newline
$\bullet$ \textbf{Caliber:} It must compute ``Gross Profit'' by aligning standalone Revenue and standalone Cost before applying the Year-over-Year (YoY) growth formula. \\
\midrule

\textbf{Input Statements} & 
A continuous time series of full financial reports: \newline
\textcolor{GtGreen}{\textbf{(1) 2021Y (Annual) Statements}} \newline
\textcolor{GtGreen}{\textbf{(2) 2022Q1 Statements}} \newline
\textcolor{GtGreen}{\textbf{(3) 2022H1 (First Half) Statements}} \newline
\textcolor{GtGreen}{\textbf{(4) 2022Q3 (First Three Quarters YTD) Statements}} \newline
\textit{(The prior-year baseline figures for 2021Q1, 2021H1, and 2021Q3 are embedded in the comparative columns of the 2022 reports.)} \\
\midrule

\textbf{User Instruction} & 
``Calculate the standalone quarterly YoY growth rates for Total Operating Revenue, Operating Profit, and Gross Profit for 2022Q1, 2022Q2, and 2022Q3. Express ratio data in percentages rounded to two decimal places. Output all final results as a JSON list wrapped in \texttt{```json} tags.'' \\
\midrule

\textbf{Injected Formula} & 
$\bullet$ \textbf{Standalone YoY Growth} = (Current Standalone Quarter $-$ Prior Year Same Standalone Quarter) / ABS(Prior Year Same Standalone Quarter) $\times 100$ \newline
$\bullet$ \textbf{Standalone Gross Profit} = Standalone Operating Revenue $-$ Standalone Operating Cost \\
\midrule

\textbf{Calculation Process} & 
\textbf{Step 1: Temporal De-cumulation (YTD to Standalone).} \newline
To find 2022Q2 standalone revenue, the model must subtract 2022Q1 revenue from 2022H1 revenue. To find 2022Q3 standalone revenue, it must subtract 2022H1 revenue from 2022Q3 YTD revenue. This process must be repeated for Operating Profit, Operating Costs, and for the 2021 base-year figures. \newline
\textbf{Step 2: Caliber Alignment (Gross Profit).} \newline
Once the standalone Q1, Q2, and Q3 Revenues and Costs are unrolled, the model subtracts Cost from Revenue to derive the standalone Gross Profit for each quarter in both 2022 and 2021. \newline
\textbf{Step 3: YoY Calculation and JSON Structuring.} \newline
Apply the YoY formula to the derived standalone metrics. Format the resulting 9 values into a 3x3 JSON array. \\
\midrule

\textbf{Ground Truth} & 
\textcolor{TagTeal}{\texttt{```json}} \newline
\textcolor{TagTeal}{\texttt{[}} \newline
\textcolor{TagTeal}{\texttt{\ \ ["}}\textcolor{GtGreen}{\textbf{11.27}}\textcolor{TagTeal}{\texttt{", "}}\textcolor{GtGreen}{\textbf{5.44}}\textcolor{TagTeal}{\texttt{", "}}\textcolor{GtGreen}{\textbf{52.17}}\textcolor{TagTeal}{\texttt{"],}} \newline
\textcolor{TagTeal}{\texttt{\ \ ["}}\textcolor{GtGreen}{\textbf{113.94}}\textcolor{TagTeal}{\texttt{", "}}\textcolor{GtGreen}{\textbf{75.23}}\textcolor{TagTeal}{\texttt{", "}}\textcolor{GtGreen}{\textbf{202.25}}\textcolor{TagTeal}{\texttt{"],}} \newline
\textcolor{TagTeal}{\texttt{\ \ ["}}\textcolor{GtGreen}{\textbf{32.07}}\textcolor{TagTeal}{\texttt{", "}}\textcolor{GtGreen}{\textbf{23.08}}\textcolor{TagTeal}{\texttt{", "}}\textcolor{GtGreen}{\textbf{58.49}}\textcolor{TagTeal}{\texttt{"]}} \newline
\textcolor{TagTeal}{\texttt{]}} \newline
\textcolor{TagTeal}{\texttt{```}} \\
\bottomrule
\end{tabularx}
\caption{\textbf{A Table-Index case testing both Temporal Reasoning and Caliber Alignment.} 
The model is provided with accumulated (YTD) financial statements and must dynamically unroll them into discrete standalone quarters before executing cross-metric calculations (Gross Profit) and temporal comparisons (YoY Growth).}
\label{tab:case_table_temporal_caliber}
\end{table*}

\paragraph{Analysis.}
This final case demonstrates the pinnacle of data-processing complexity in \textsc{FinIndices}, combining all three primary friction points: structural output generation (\textcolor{IdxBlue}{Table-Index}), \textcolor{TagTeal}{Temporal Reasoning}, and \textcolor{TagTeal}{Caliber Alignment}. 

In standard tabular QA benchmarks, calculating a Year-over-Year (YoY) metric simply requires locating a ``Current Period'' column and a ``Prior Period'' column in the same table. In real-world financial analysis, however, interim reports (such as Q3) report \emph{cumulative Year-to-Date} (YTD) figures. When tasked with analyzing a \emph{standalone} quarter (e.g., Q2 or Q3 specifically), an analyst must retrieve multiple historical reports and subtract the previous cumulative figures from the current ones. 

Here, the LLM must first temporally de-cumulate the flows for Revenue, Profit, and Cost across multiple years. Once the discrete standalone figures are isolated, it must align calibers by calculating Gross Profit (Revenue minus Cost) strictly within the bounds of those isolated quarters. A single extraction error, a failure to subtract the H1 figures from the Q3 YTD figures, or an inability to maintain the JSON matrix structure will result in cascading failures, perfectly mirroring the strict reliability demands of industrial financial agents.

\section{Detailed Results across Capability Axes}
\label{sec:appendix_detailed_results}

To provide a more granular understanding of model performance, we break down the Test set accuracy of the leading general Large Language Models across the four fine-grained capability axes: \textit{Domain Understanding}, \textit{Complex Caliber Alignment}, \textit{Temporal Reasoning + Complex Caliber Alignment}, and \textit{Temporal Reasoning + Domain Understanding}. 

Table~\ref{tab:detailed_with_hint} presents the detailed results when calculation formulas are explicitly provided (With Hint), and Table~\ref{tab:detailed_without_hint} presents the results when formulas are withheld (Without Hint). Note that all evaluated general models were run in their explicit reasoning (thinking) modes.

\begin{table*}[h]
\centering
\small
\setlength{\tabcolsep}{4pt}
\renewcommand{\arraystretch}{1.2}
\resizebox{\textwidth}{!}{
\begin{tabular}{@{} l cccc cc c @{}}
\toprule
\textbf{Capability Axis} & \textbf{Claude-4.5} & \textbf{Gemini-3.1} & \textbf{GPT-5.4} & \textbf{DS-V4-Pro} & \textbf{GLM-5} & \textbf{Qwen3.5-Plus} & \textbf{Qwen3.5-35B-A3B} \\
\midrule
\multicolumn{8}{@{}l}{\textit{\textbf{Task: Single-Index}}} \\
\midrule
Domain Understanding         & 78.79 & 76.47 & 71.88 & 71.88 & 61.76 & 69.70 & 61.29 \\
Caliber Alignment            & 64.00 & 84.90 & 81.44 & 67.91 & 54.13 & 72.92 & 67.43 \\
Temporal + Caliber           & 59.42 & 74.67 & 68.12 & 59.26 & 46.25 & 69.57 & 61.54 \\
Temporal + Domain            & 52.83 & 59.26 & 58.33 & 34.55 & 32.73 & 42.55 & 47.27 \\
\midrule
\multicolumn{8}{@{}l}{\textit{\textbf{Task: Table-Index}}} \\
\midrule
Domain Understanding         & 63.64 & 63.64 & 72.73 & 68.18 & 45.45 & 68.18 & 27.27 \\
Caliber Alignment            & 27.91 & 63.95 & 50.00 & 32.56 & 20.93 & 52.33 & 23.26 \\
Temporal + Caliber           & 56.25 & 90.62 & 90.62 & 59.38 & 37.50 & 78.12 & 40.62 \\
Temporal + Domain            & 76.47 & 76.47 & 88.24 & 76.47 & 88.24 & 88.24 & 29.41 \\
\bottomrule
\end{tabular}
}
\caption{\textbf{Detailed Test Accuracy (\%) on \textsc{FinIndices} (With Hint).} Results are grouped by task paradigm and capability axes.}
\label{tab:detailed_with_hint}
\end{table*}

\begin{table*}[h]
\centering
\small
\setlength{\tabcolsep}{4pt}
\renewcommand{\arraystretch}{1.2}
\resizebox{\textwidth}{!}{
\begin{tabular}{@{} l cccc cc c @{}}
\toprule
\textbf{Capability Axis} & \textbf{Claude-4.5} & \textbf{Gemini-3.1} & \textbf{GPT-5.4} & \textbf{DS-V4-Pro} & \textbf{GLM-5} & \textbf{Qwen3.5-Plus} & \textbf{Qwen3.5-35B-A3B} \\
\midrule
\multicolumn{8}{@{}l}{\textit{\textbf{Task: Single-Index}}} \\
\midrule
Domain Understanding         & 55.88 & 62.50 & 48.48 & 42.42 & 37.50 & 43.75 & 48.48 \\
Caliber Alignment            & 51.46 & 68.47 & 61.05 & 48.98 & 37.91 & 54.26 & 50.17 \\
Temporal + Caliber           & 54.41 & 60.81 & 58.73 & 51.85 & 41.46 & 44.12 & 45.00 \\
Temporal + Domain            & 54.72 & 51.92 & 38.78 & 35.85 & 25.00 & 40.00 & 50.00 \\
\midrule
\multicolumn{8}{@{}l}{\textit{\textbf{Task: Table-Index}}} \\
\midrule
Domain Understanding         & 18.18 & 22.73 & 18.18 & 27.27 &  9.09 & 18.18 &  4.55 \\
Caliber Alignment            & 12.79 & 23.26 & 15.12 & 11.63 &  9.30 & 13.95 &  9.30 \\
Temporal + Caliber           & 37.50 & 68.75 & 59.38 & 50.00 & 25.00 & 56.25 & 15.62 \\
Temporal + Domain            & 76.47 & 76.47 & 76.47 & 58.82 & 70.59 & 47.06 & 41.18 \\
\bottomrule
\end{tabular}
}
\caption{\textbf{Detailed Test Accuracy (\%) on \textsc{FinIndices} (Without Hint).} The removal of explicit formulas causes a systemic collapse across all capability axes, particularly on pure domain and caliber alignment tasks.}
\label{tab:detailed_without_hint}
\end{table*}

Several key observations can be drawn from this fine-grained breakdown:

\paragraph{The ``Temporal + Domain'' Challenge.} 
In the \textit{Single-Index} setting, almost all models exhibit their worst performance on the \textit{Temporal + Domain Understanding} axis (e.g., GPT-5.4 scores $58.33\%$ with hints, dropping to $38.78\%$ without hints). This axis often requires Trailing Twelve Months (TTM) calculations, forcing models to temporally roll across 4--5 non-sequential historical reports while concurrently applying textbook accounting taxonomy (e.g., netting out specific non-interest-bearing items). The compounding complexity of multi-hop extraction and deep domain classification remains a massive hurdle.

\paragraph{Collapse in ``Complex Caliber Alignment'' during Table Generation.} 
In the \textit{Table-Index} setting, \textit{Complex Caliber Alignment} emerges as the most severely bottlenecked capability. Even with formulas provided (Table~\ref{tab:detailed_with_hint}), the leading model (Gemini-3.1) only achieves $63.95\%$, and GPT-5.4 drops to $50.00\%$. This task often involves unifying point-in-time stock metrics (e.g., end-of-period inventory) with period-aggregated flow metrics (e.g., annual revenue) to compute turnover rates across multiple consecutive years. When models must maintain this strict stock-flow reconciliation simultaneously across multiple cells in a 2D matrix, structural hallucinations and alignment drift become highly prevalent.

\paragraph{Extreme Vulnerability Without Domain Hints.} 
Comparing Table~\ref{tab:detailed_with_hint} and Table~\ref{tab:detailed_without_hint}, the absence of explicit formulas disproportionately devastates \textit{Domain Understanding} and \textit{Caliber Alignment} on the Table-Index task. For instance, GPT-5.4's accuracy on Domain Understanding collapses from $72.73\%$ to $18.18\%$, and Gemini drops from $63.95\%$ to $23.26\%$ on Caliber Alignment. This definitively proves that current LLMs lack intrinsic memorization of standard financial formulas. Without external hints guiding them, models routinely default to simplified, statistically flawed estimations (e.g., ignoring minority interests or failing to average beginning/ending balances), rendering them unsuitable for autonomous, high-stakes financial auditing.


\section{Weakness Study}
\label{sec:weakness_study}

\subsection{Anomaly on the Leaderboard: Why Do Large-Scale or Closed-Source Models with Strong General Capabilities Fail?}
\label{subsec:why_large_models_fail}

On our evaluation leaderboard, several large-scale or closed-source models with formidable general capabilities exhibit anomalously poor performance compared to their peers on complex financial tasks. To systematically investigate this performance bottleneck within \textsc{FinIndices}, we conducted a highly targeted qualitative analysis. By meticulously tracing generation trajectories—and contrasting model behavior between \texttt{single\_indice} (isolated extraction) and \texttt{table\_indice} (structured table) tasks—we uncovered the fundamental reasons behind their collapse in end-to-end financial reasoning. 

Our analysis reveals that these failures predominantly stem from profound knowledge bottlenecks across our Capability Axes and a severe vulnerability to \textit{rigid formula disobedience} under structural constraints. Rather than activating an internal financial knowledge base, these models often rely on \textit{semantic surface-matching}. They frequently fall into semantic anchor traps, suffer from cognitive overload during multi-hop temporal de-cumulation, ignore the physical mismatch between point-in-time stocks and period flows, and fail to distinguish fine-grained accounting taxonomies.

To summarize and conceptualize these flaws, we selected \textbf{5 highly representative cases} spanning diverse failure modes. Table~\ref{tab:consolidated_large_model_cases} illustrates how the models' erroneous trajectories deviate from correct domain logic.

\begin{table*}[htbp]
\centering
\small
\renewcommand{\arraystretch}{1.6}
\begin{tabularx}{\textwidth}{@{} >{\hsize=0.35\hsize\raggedright\arraybackslash\bfseries}X >{\hsize=1.65\hsize\raggedright\arraybackslash}X @{}}
\toprule

\rowcolor[gray]{0.9}
\multicolumn{2}{c}{\textbf{Case 1: Semantic Anchor Trap (Domain Axis)}} \\ 
Target \& Context & Net Income from Value Changes $\vert$ 2025H1 Income Statement (RuiDi Drives). \\
Ground Truth & \textcolor{green!50!black}{\textbf{289.1}} (10k RMB) \hspace{0.2cm} $\vert$ \hspace{0.2cm} \textit{Formula:} Investment Income + Fair Value Change + Exchange Gain + Hedging Gain. \\
Erroneous Trajectory & \textcolor{red}{\textbf{-24.5}}. Triggered by the literal string ``Value Change,'' the model solely extracted ``Fair value change income'' (-245,236) and ignored the ``Investment Income'' (3,136,526) necessary for the full aggregated formula. \\
Root Cause & \textbf{Semantic Rigidity.} The model lacks the internal domain mapping to recognize that ``Value Change'' is an umbrella financial indicator, falling for a surface-level lexical match. \\ \midrule

\rowcolor[gray]{0.9}
\multicolumn{2}{c}{\textbf{Case 2: Cognitive Overload in De-cumulation (Temporal Axis)}} \\ 
Target \& Context & 2022Q3 Single-Quarter Gross Profit YoY Growth $\vert$ 2022Q3 YTD, 2022H1, and 2021 statements (Zhongzhou). \\
Ground Truth & \textcolor{green!50!black}{\textbf{58.49\%}} \hspace{0.2cm} $\vert$ \hspace{0.2cm} \textit{Formula:} $GP_{Q3\_single} = (Rev_{YTD} - Rev_{H1}) - (Cost_{YTD} - Cost_{H1})$. \\
Erroneous Trajectory & \textcolor{red}{\textbf{202.25\%}}. The model correctly computed the 2022Q3 single-quarter profit. However, for the baseline year (2021Q3), it lost track of the constraint and directly used the 2021Q3 \textit{YTD} profit as the denominator. \\
Root Cause & \textbf{Contextual Amnesia.} When nesting temporal subtractions inside a comparative ratio formula, the attention mechanism fails to replicate the de-cumulation logic across historical reporting years. \\ \midrule

\rowcolor[gray]{0.9}
\multicolumn{2}{c}{\textbf{Case 3: Ignoring Structural Caliber (Caliber Axis)}} \\ 
Target \& Context & Annualized Return on Assets (ROA) $\vert$ 2025Q3 Income Statement, 2025Q3 \& 2024Y Balance Sheets (Suneng). \\
Ground Truth & \textcolor{green!50!black}{\textbf{1.217\%}} \hspace{0.2cm} $\vert$ \hspace{0.2cm} \textit{Formula:} $(Net Profit_{Q3} \times \frac{4}{3}) / Average Total Assets$. \\
Erroneous Trajectory & \textcolor{red}{\textbf{1.208\%}}. The model successfully annualized the profit flow but directly divided it by the \textit{Ending Total Assets} (39.74B), completely ignoring the Beginning Assets. \\
Root Cause & \textbf{Stock-Flow Mismatch.} The model treats numerical extractions as flat variables, demonstrating a severe lack of caliber alignment awareness (averaging stocks to match flows) necessary for cross-statement integration. \\ \midrule

\rowcolor[gray]{0.9}
\multicolumn{2}{c}{\textbf{Case 4: Rigid Formula Disobedience}} \\ 
Target \& Context & Cash Return on Investment Ratio $\vert$ Xiangtan Elec. 2018Q3 (Round to 1 decimal). \\
Ground Truth & \textcolor{green!50!black}{\textbf{0.5}} \hspace{0.2cm} $\vert$ \hspace{0.2cm} \textit{Formula:} Net Operating CF / (Fixed Asset Cash Paid + Dividends - Financial Exp.) \\
Erroneous Trajectory & \textcolor{red}{\textbf{0.4}}. The model computed the denominator as 386,026,411.74. \\
Root Cause & \textbf{Instruction Disobedience.} The model failed to strictly follow the multi-step denominator formula provided in the prompt. It selectively omitted the required subtraction of ``Financial Expenses'', showing rigidity when facing custom composite indicators. \\ \midrule

\rowcolor[gray]{0.9}
\multicolumn{2}{c}{\textbf{Case 5: Conceptual Misalignment and Taxonomy Confusion}} \\ 
Target \& Context & Tangible Asset Net Value $\vert$ Zhixin Precision 2025H1 (Unit: 10M RMB). \\
Ground Truth & \textcolor{green!50!black}{\textbf{97.0}} \hspace{0.2cm} $\vert$ \hspace{0.2cm} \textit{Formula:} Equity Attributable to Parent - (Intangibles + Goodwill + Deferred Tax Assets, etc.) \\
Erroneous Trajectory & \textcolor{red}{\textbf{99.8}}. The model used generic ``Total Owner's Equity'' as the base. \\
Root Cause & \textbf{Taxonomy Confusion.} The model failed to distinguish the fine-grained boundary between ``Total Equity'' and ``Equity Attributable to Parent Company''. Furthermore, it missed deductions required by new accounting standards (e.g., Deferred Tax Assets). \\ \bottomrule

\end{tabularx}
\caption{Taxonomy of 5 representative failure modes for large-scale models. By contrasting flawed generation trajectories against correct domain logic, we expose severe vulnerabilities in temporal de-cumulation, caliber alignment, and formula adherence.}
\label{tab:consolidated_large_model_cases}
\end{table*}

\subsection{Small Parameter Models (e.g., 35B variants): Magnitude Scaling and Format Brittleness}
\label{subsec:flash_additional_weakness}

Beyond cognitive bottlenecks, we identified a critical arithmetic vulnerability in underperforming small parameter models (e.g., roughly 35B variants): \textbf{Magnitude Scaling Brittleness}. This error, often combined with a structural \textbf{Helpfulness Bias}, explains the performance collapse in \texttt{table\_indice} tasks. Table~\ref{tab:case_3_format_scaling} illustrates a typical case where correct extraction logic is ruined by faulty unit conversion and formatting hallucination.

\begin{table*}[htbp]
\centering
\small
\renewcommand{\arraystretch}{1.6}
\begin{tabularx}{\textwidth}{@{} >{\hsize=0.4\hsize\raggedright\arraybackslash\bfseries}X >{\hsize=1.6\hsize\raggedright\arraybackslash}X @{}}
\toprule
\rowcolor[gray]{0.9}
\multicolumn{2}{c}{\textbf{Small Model Case: Scaling Collapse and Format Hallucination}} \\ \midrule
\textbf{Target Metric} & Working Capital, Retained Earnings, Total Invested Capital (2022Y - 2025Q3) \\
\textbf{Constraints} & 1. Unit: 10,000 RMB (Divide raw values by $10^4$). \newline
2. Format: \texttt{tuple\_list} (e.g., \texttt{[["Year", Val1, Val2, ...]]}). \newline
3. Precision: Round to 2 decimal places. Frequency: Annual basis. \\ 
\textbf{Ground Truth} & \textcolor{green!50!black}{\texttt{[["2022Y", -19368.16, 75991.92, 274889.0], ...]]}} \newline \textit{(Strict division by $10^4$, correct formula logic, compliant tuple formatting)} \\ 
\textbf{Model Output} & \textcolor{red}{\texttt{[\{"Year": "2022Y", "Working\_Capital\_10k": -1936.82, "Retained\_Earnings\_10k": 7599.19, "Total\_Invested\_Capital\_10k": 27494.20\}, ...\}]}} \\ \bottomrule
\end{tabularx}
\caption{Scaling collapse and format hallucination in small parameter models. The model fails the arithmetic scaling step (incorrect decimal shift) and hallucinates JSON dictionary keys instead of the requested nested tuple array.}
\label{tab:case_3_format_scaling}
\end{table*}

\paragraph{Failure Analysis:}
Two primary mechanisms drive this degradation in smaller models:
\begin{enumerate}
    \item \textbf{Magnitude Scaling Errors:} Professional financial tasks often require converting raw monetary values to 10k RMB (division by $10^4$) or Millions. As shown in Table~\ref{tab:case_3_format_scaling}, the model successfully identifies the correct intermediate values but fails the scalar translation step, shifting the decimal point incorrectly (e.g., yielding $-1936.82$ instead of $-19368.16$). 
    \item \textbf{Helpfulness Bias:} The model frequently violates the \texttt{tuple\_list} constraint by hallucinating descriptive dictionary keys (e.g., \texttt{"Working\_Capital\_10k"}). This unsolicited over-optimization intended to increase human-readability breaks the programmatic JSON schema, leading to automatic parsing failures.
\end{enumerate}

\section{Analysis of the ``Structure Bottleneck''}
\label{appendix:structure_bottleneck}

In our evaluation of ultra-long context financial reasoning, we identified a profound failure mode termed the \textbf{Structure Bottleneck}. When queried for a single financial metric in isolation (\texttt{single\_indice}), modern LLMs act as expert financial agents, flawlessly executing complex multi-period temporal logic and deep accounting derivations. However, when required to generate multi-metric, multi-period structured outputs (e.g., nested JSONs or tabular matrices in \texttt{table\_indice} tasks), the cognitive load of adhering to rigid structural schemas induces severe reasoning drift. 

To demonstrate that this is a format-induced cognitive collapse rather than a fundamental lack of financial knowledge, we conducted an \textbf{indicator-level controlled ablation study}. By contrasting the model's performance on isolated metrics against identical indicator types embedded within batch-generation tasks, we observe that the Structure Bottleneck degrades model reasoning in two primary dimensions: \textbf{Temporal Misalignment} and \textbf{Aggregation Shortcut}.

\subsection{Case 1: Temporal Misalignment in Time-Series Logic}

The first manifestation of the Structure Bottleneck is the loss of temporal anchoring. Complex time-series calculations, such as Year-over-Year (YoY) growth rates, require precise mapping of cross-period financial reports.

Table~\ref{tab:structure_bottleneck_temporal} isolates the indicator types requested in the \textit{Tongdahai} multi-metric table task. When evaluating the YoY growth of Net Assets in isolation (e.g., \textit{Xinlaifu}), the model accurately identifies that it must locate the ``Prior Year Same Period'' (e.g., Q1 of the previous year) rather than settling for the ``Beginning of Year'' balance. 

However, when forced to populate a comprehensive HTML table tracking four different YoY metrics for \textit{Tongdahai}, the model's temporal rigor collapses. Because standard Balance Sheets only display ``End of Period'' and ``Beginning of Period'' (without an explicit ``Prior Year Same Period'' column), finding the correct denominator requires cross-report retrieval. Overwhelmed by the table generation constraint, the model takes a severe cognitive shortcut: it lazily grabs the adjacent ``Beginning of Period'' value to force-fill the table cell, treating it as the prior year's value. This structural coercion corrupts the temporal logic, transforming the ground truth ($-4.22\%$) into a hallucinated mathematical artifact ($-3.25\%$).

\begin{table*}[htbp]
\centering
\scriptsize
\renewcommand{\arraystretch}{1.5}
\begin{tabularx}{\textwidth}{@{} 
>{\hsize=0.80\hsize\raggedright\arraybackslash\bfseries}X 
>{\hsize=1.10\hsize\raggedright\arraybackslash}X 
>{\hsize=1.00\hsize\raggedright\arraybackslash}X 
>{\hsize=1.00\hsize\raggedright\arraybackslash}X 
>{\hsize=1.10\hsize\raggedright\arraybackslash}X 
@{}}
\toprule
\rowcolor{blue!10}
\textcolor{blue!50!black}{\textbf{Financial Indicator}} & 
\textcolor{blue!50!black}{\textbf{Accounting \& Calculation Logic}} & 
\textcolor{blue!50!black}{\textbf{Isolated Query (\texttt{single\_indice})}} & 
\textcolor{blue!50!black}{\textbf{Table Query (\texttt{table\_indice})}} & 
\textcolor{blue!50!black}{\textbf{Bottleneck Analysis}} \\ \midrule

\rowcolor{red!5}
\textbf{Net Asset YoY Growth} & 
\textbf{Temporal mapping: $\frac{\text{Current End} - \text{Prior Year Same Period}}{\text{ABS(Prior Year Same Period)}}$} & 
\textbf{\textit{Xinlaifu:}} \mycmark \newline \textbf{\textcolor{green!50!black}{GT: 6.19\% \newline Pred: 6.19\%}} & 
\textbf{\textit{Tongdahai:}} \myxmark \newline \textbf{\textcolor{red}{GT: -4.22\% \newline Pred: -3.25\%}} & 
\textbf{Temporal Misalignment (Column Substitution).} \newline In isolation, the model correctly locates the Prior Year Same Period report. In the table, it lazily substitutes the adjacent ``Beginning Balance'' column to force-fill the structural schema: $(1311.76M - 1355.79M) / 1355.79M = -3.25\%$. \\ \midrule

Net Profit Cash Content YoY & 
Complex cross-statement derivation: Compute Operating CF / Net Profit, then calculate YoY. & 
\textit{Gelin Jingmi:} \mycmark \newline \textcolor{gray}{GT: 12349.1\% \newline Pred: 12349.1\%} & 
\textit{Tongdahai:} \mycmark \newline \textcolor{gray}{GT: -11.48\% \newline Pred: -11.48\%} & 
Maintained. The model can execute complex cross-statement math when the required YoY columns are explicitly aligned in the source statements. \\ 

Operating Profit YoY & 
Standard line-item YoY derivation: Deduct prior year operating profit; divide by ABS prior year. & 
\textit{Zhongzhou Tecai:} \mycmark \newline \textcolor{gray}{GT: 6.0\% \newline Pred: 6.0\%} & 
\textit{Tongdahai:} \mycmark \newline \textcolor{gray}{GT: -29.34\% \newline Pred: -29.34\%} & 
Maintained. Simple horizontal math extraction survives the structural load because the ``Prior Period'' column is explicitly provided in the Income Statement. \\ \bottomrule
\end{tabularx}
\caption{\centering Ablation of Temporal Misalignment. By evaluating identical YoY indicator types individually versus inside a unified HTML table, we isolate the failure trigger. Gemini possesses the temporal logic to fetch correct prior-period reports (evidenced by 100\% isolated accuracy), but systematically substitutes wrong, adjacent columns (e.g., Beginning Balance) to lazily fulfill structural constraints under the cognitive load of multi-metric table generation.}
\label{tab:structure_bottleneck_temporal}
\end{table*}

\subsection{Case 2: Aggregation Shortcut in Accounting Logic}

The second variant manifests as an ``Aggregation Shortcut.'' To sustain the generation of a long, multi-metric list, the model abandons deep semantic accounting reasoning and falls back on shallow, top-level aggregate parsing.

Table~\ref{tab:structure_bottleneck_aggregation} isolates the exact five indicator types requested in the \textit{Zhenyu Tech} multi-metric table task. When queried individually, Gemini acts flawlessly, executing complex cross-period deductions and nuanced accounting derivations. Crucially, it demonstrates perfect understanding of the ``Operating Net Income / Total Profit'' formula in isolation (scoring 104.0\% for \textit{ST Tongde}). 

However, when these identical cognitive tasks are bundled into a unified five-metric table, the model's reasoning fractures. While it successfully maintains rigor for simpler ratios, it suffers a catastrophic collapse specifically on the most accounting-sensitive metric. Overwhelmed by the tuple-list constraint, the model defaults to a literal $(Revenue - Cost) / Profit$ shortcut ($129.82\%$), completely ignoring Chinese Accounting Standards (CAS) impairment line items that it successfully processed during the isolated \texttt{single\_indice} query.

\begin{table*}[htbp]
\centering
\scriptsize
\renewcommand{\arraystretch}{1.5}
\begin{tabularx}{\textwidth}{@{} 
>{\hsize=0.80\hsize\raggedright\arraybackslash\bfseries}X 
>{\hsize=1.10\hsize\raggedright\arraybackslash}X 
>{\hsize=1.00\hsize\raggedright\arraybackslash}X 
>{\hsize=1.00\hsize\raggedright\arraybackslash}X 
>{\hsize=1.10\hsize\raggedright\arraybackslash}X 
@{}}
\toprule
\rowcolor{blue!10}
\textcolor{blue!50!black}{\textbf{Financial Indicator}} & 
\textcolor{blue!50!black}{\textbf{Accounting \& Calculation Logic}} & 
\textcolor{blue!50!black}{\textbf{Isolated Query (\texttt{single\_indice})}} & 
\textcolor{blue!50!black}{\textbf{Table Query (\texttt{table\_indice})}} & 
\textcolor{blue!50!black}{\textbf{Bottleneck Analysis}} \\ \midrule

\rowcolor{red!5}
\textbf{Operating Net / Total Profit} & 
\textbf{Deep CAS adjustment (incorporating Impairment Losses outside OpCost).} & 
\textbf{\textit{ST Tongde:}} \mycmark \newline \textbf{\textcolor{green!50!black}{GT: 104.0\% \newline Pred: 104.0\%}} & 
\textbf{\textit{Zhenyu Tech:}} \myxmark \newline \textbf{\textcolor{red}{GT: 94.06\% \newline Pred: 129.82\%}} & 
\textbf{Aggregation Shortcut.} In isolation, Gemini executes the CAS logic perfectly. Inside the table, it defaults to a lazy literal $(Rev-Cost)/Profit$ shortcut to save cognitive effort. \\ \midrule

Single-Quarter ROE & 
Deduct prior quarter net profit; divide by average parent equity. & 
\textit{Dingrongyan:} \mycmark \newline \textcolor{gray}{GT: 4.6\% \newline Pred: 4.6\%} & 
\textit{Zhenyu Tech:} \mycmark \newline \textcolor{gray}{GT: 3.70\% \newline Pred: 3.70\%} & 
Maintained. Standard ratio logic survives the structural load. \\ 

Admin Expense / Revenue Ratio & 
Deduct prior quarter values to isolate single quarter; compute ratio. & 
\textit{Zhongjin:} \mycmark \newline \textcolor{gray}{GT: 18.2\% \newline Pred: 18.2\%} & 
\textit{Zhenyu Tech:} \mycmark \newline \textcolor{gray}{GT: 2.73\% \newline Pred: 2.73\%} & 
Maintained. Shallow numerator-denominator mapping remains stable. \\ 

Value-Change Net Income & 
Sum fair-value change, investment income, and FX income; adjust unit. & 
\textit{Zanyu Tech:} \mycmark \newline \textcolor{gray}{GT: 13,333.95 \newline Pred: 13,333.95} & 
\textit{Zhenyu Tech:} \mycmark \newline \textcolor{gray}{GT: 139,014.56 \newline Pred: 139,014.56} & 
Maintained. Additive aggregation remains unaffected by table constraints. \\ 

Investing Cash Flow Ratio & 
Aggregate three distinct CF categories; compute proportional share. & 
\textit{Tianxin:} \mycmark \newline \textcolor{gray}{GT: -97.15\% \newline Pred: -97.15\%} & 
\textit{Zhenyu Tech:} \mycmark \newline \textcolor{gray}{GT: -176.44\% \newline Pred: -176.44\%} & 
Maintained. Multi-step but structurally salient on the cash flow statement. \\ \bottomrule
\end{tabularx}
\caption{\centering Ablation of the Aggregation Shortcut. By evaluating the \textit{exact same five indicators} individually versus as a unified table, we isolate the failure trigger. Gemini possesses the knowledge to solve complex operating adjustments (evidenced by 100\% isolated accuracy), but systematically abandons this rigor for a shallow shortcut under the cognitive load of multi-metric table generation.}
\label{tab:structure_bottleneck_aggregation}
\end{table*}

Ultimately, the Structure Bottleneck reveals a zero-sum cognitive mechanism within the LLM: allocating internal tokens and attention to maintain rigid multi-dimensional JSON matrices, HTML tables, or Tuple Lists actively drains the reasoning resources required for deep financial parsing and multi-step temporal math.

\section{Source Materials}
\label{appendix:source_materials}

\subsection{Formula Source Reference}
\label{appendix:formula_ref}
The integrity and precision of financial index calculations are fundamentally predicated on the conceptual validity of their underlying formulas. To ensure that the \textbf{FinIndices} benchmark encapsulates professional-grade financial reasoning and analytical depth, we have meticulously mapped over 300 indices to a comprehensive spectrum of authoritative sources. These foundations span from rigorous statutory accounting standards used in global regulatory filings to seminal textbooks and frameworks central to the CFA and CPA curricula, providing a robust ground truth for multi-dimensional financial evaluation. Detailed source distributions and representative formulaic logic are provided in Table~\ref{tab:formula_sources}.

\definecolor{tablegreen}{rgb}{0.88, 0.95, 0.88}    
\definecolor{darkgreen}{rgb}{0.0, 0.35, 0.0}      
\definecolor{varred}{rgb}{0.8, 0.0, 0.0}
\definecolor{varblue}{rgb}{0.0, 0.0, 0.8}
\definecolor{vargreen}{rgb}{0.0, 0.5, 0.0}

\begin{table*}[htbp]
\centering
\small
\renewcommand{\arraystretch}{2.0}
\begin{tabularx}{\textwidth}{@{} >{\hsize=0.45\hsize\raggedright\arraybackslash\bfseries}X >{\hsize=0.15\hsize\centering\arraybackslash}X >{\hsize=0.8\hsize\raggedright\arraybackslash}X >{\hsize=1.6\hsize\raggedright\arraybackslash}X @{}}
\toprule
\rowcolor{tablegreen}
\textcolor{darkgreen}{\textbf{Source Identifier}} & \textcolor{darkgreen}{\textbf{Ratio}} & \textcolor{darkgreen}{\textbf{Source Overview}} & \textcolor{darkgreen}{\textbf{Representative Examples}} \\ \midrule

Financial Reporting Analysis (Revsine) & 58.05\% & An analytical framework utilized for global financial statement analysis. It establishes the methodology for period-to-period reconciliation and Trailing Twelve Months (TTM) adjustments to ensure cross-period comparability. & 
\textbullet\ \texttt{\textcolor{varred}{OCF\_Margin}(TTM) = \textcolor{varblue}{OCF(TTM)} / \textcolor{vargreen}{Revenue(TTM)}} \newline 
\textbullet\ \texttt{\textcolor{varred}{ROIC}(TTM) = \textcolor{varblue}{NOPAT(TTM)} / \textcolor{vargreen}{Avg\_Invested\_Capital}} \newline 
\textbullet\ \texttt{\textcolor{varred}{Net\_Profit\_Ratio}(TTM) = \textcolor{varblue}{Net\_Profit(TTM)} / \textcolor{vargreen}{Revenue(TTM)}} \\ \midrule

Chinese Accounting Standards (CAS) & 35.34\% & The statutory reporting standard issued by the Ministry of Finance of China. It defines the rigorous line-item classification for listed companies across general and financial (Banking, Securities, Insurance) sectors. & 
\textbullet\ \texttt{\textcolor{varred}{Interest\_Bearing\_Debt} = \textcolor{varblue}{Total\_Liabs} - \textcolor{vargreen}{Non-interest\_Liabs}} \newline 
\textbullet\ \texttt{\textcolor{varred}{EBITDA} = \textcolor{varblue}{Op\_Profit} + \textcolor{vargreen}{Depr\_Amort} + \textcolor{varblue}{Interest\_Exp}} \newline 
\textbullet\ \texttt{\textcolor{varred}{Operating\_Revenue} = \textcolor{varblue}{Net\_Interest} + \textcolor{varblue}{Net\_Fee} + \dots} \\ \midrule

\textbf{Textbooks} & 6.61\% & Derived from authoritative global textbooks, including McKinsey's \textit{Valuation}, Ross's \textit{Corporate Finance}, and Kieso's \textit{Intermediate Accounting}, which serve as standard academic references. & 
\textbullet\ \textbf{FCFF:} \texttt{\textcolor{varred}{EBIT(1-t)} + \textcolor{varblue}{D\&A} - \textcolor{vargreen}{CapEx} - \textcolor{vargreen}{$\Delta$Working\_Cap}} \newline 
\textbullet\ \textbf{Cash Cycle:} \texttt{\textcolor{varblue}{Inv\_Days} + \textcolor{varblue}{Rec\_Days} - \textcolor{vargreen}{Pay\_Days}} \newline 
\textbullet\ \textbf{ROE:} \texttt{\textcolor{varred}{Net\_Income} / \textcolor{vargreen}{[(Equity\_beg + Equity\_end)/2]}} \\ \bottomrule
\end{tabularx}
\caption{\textbf{Comprehensive Taxonomy of Formulaic Sources}. The ``Ratio'' denotes the percentage of total formulas derived from each framework. Color coding: \textcolor{varred}{Target Ratio/Metric}, \textcolor{varblue}{Core Financial Item}, and \textcolor{vargreen}{Adjustment/Normalization Factor}.}
\label{tab:formula_sources}
\end{table*}

\end{document}